\documentclass[sn-mathphys,Numbered]{sn-jnl}% Math and Physical Sciences Reference Style

\usepackage{graphicx}%
\usepackage{multirow}%
\usepackage{amsmath,amssymb,amsfonts}%
\usepackage{amsthm}%
\usepackage{amsmath}
\usepackage{mathrsfs}%
\usepackage[title]{appendix}%
\usepackage{xcolor}%
\usepackage{textcomp}%
\usepackage{manyfoot}%
\usepackage{booktabs}%
\usepackage{algorithm}%
\usepackage{algorithmicx}%
\usepackage{algpseudocode}%
\usepackage{listings}%
\usepackage{adjustbox}
\usepackage{lineno}
\usepackage{enumitem}
\usepackage{arabtex}
\usepackage{utf8}
\setcode{utf8}
\def\endabstract{\egroup}
\usepackage{graphicx}
\usepackage{amsmath}
\usepackage{algorithmicx}
\usepackage{algorithm}
\usepackage{algpseudocode}
\usepackage{algcompatible}
\usepackage{caption}
\usepackage{booktabs}
\usepackage{multirow}
\usepackage{longtable}
\usepackage{supertabular,booktabs}
\usepackage{subcaption}
\usepackage{bigstrut}
\usepackage{colortbl}
\usepackage{marvosym}
\usepackage[english]{babel}
\usepackage[utf8x]{inputenc}
\usepackage[T1]{fontenc}
\usepackage{float}
\usepackage{enumitem}
\usepackage{array}
\usepackage{afterpage}
\usepackage{pdflscape}
\usepackage{arabtex}
\usepackage{utf8}
\usepackage{setspace} 
\usepackage{esvect}
\usepackage{fancyvrb}
\usepackage{breqn}
\usepackage{savesym}
\usepackage{xcolor}
\usepackage{graphicx}
\usepackage{booktabs}
\DeclareUnicodeCharacter{2212}{\ensuremath{-}}
\algnewcommand\algorithmicinput{\textbf{Input:}}
\algnewcommand\INPUT{\item[\algorithmicinput]}
\algnewcommand\algorithmicoutput{\textbf{Output:}}
\algnewcommand\OUTPUT{\item[\algorithmicoutput]}

\begin{document}

\title[Article Title]{An Ensemble Model with Attention Based Mechanism for Image Captioning}


\author[1]{\fnm{Israa} \sur{Al Badarneh}}%\email{ASR9170348@ju.edu.jo}

\author[1,2]{\fnm{Bassam H.} \sur{Hammo}}%\email{b.hammo@ju.edu.jo}

\author[1]{\fnm{Omar} \sur{Al-Kadi}}%\email{o.alkadi@ju.edu.jo}

\affil[1]{\orgdiv{King Abdullah II School for Information Technology}, \orgname{The University of Jordan}, \orgaddress{
\city{Amman},
\country{Jordan}}}

\affil[2]{\orgdiv{King Hussein School of Computing Sciences}, \orgname{Princess Sumaya University for Technology},
\orgaddress{ 
\city{Amman},
\country{Jordan}}}



\abstract{Image captioning creates informative text from an input image by creating a relationship between the words and the actual content of an image. Recently, deep learning models that utilize transformers have been the most successful in automatically generating image captions. The capabilities of transformer networks have led to notable progress in several activities related to vision. In this paper, we thoroughly examine transformer models, emphasizing the critical role that attention mechanisms play. The proposed model uses a transformer encoder-decoder architecture to create textual captions and a deep learning convolutional neural network to extract features from the images. To create the captions, we present a novel ensemble learning framework that improves the richness of the generated captions by utilizing several deep neural network architectures based on a voting mechanism that chooses the caption with the highest bilingual evaluation understudy (BLEU) score. The proposed model was evaluated using publicly available datasets. Using the Flickr8K dataset, the proposed model achieved the highest BLEU-[1-3] scores with rates of 0.728, 0.495, and 0.323, respectively. The suggested model outperformed the latest methods in Flickr30k datasets, determined by BLEU-[1-4] scores with rates of 0.798, 0.561, 0.387, and 0.269, respectively. The model efficacy was also obtained by the Semantic propositional image caption evaluation (SPICE) metric with a scoring rate of 0.164 for the Flicker8k dataset and 0.387 for the Flicker30k. Finally, ensemble learning significantly advances the process of image captioning and, hence, can be leveraged in various applications across different domains.}




\maketitle

\section{Introduction}
\begin{figure*}
    \centering
    \includegraphics[width=0.9\textwidth,height=\textheight,keepaspectratio]{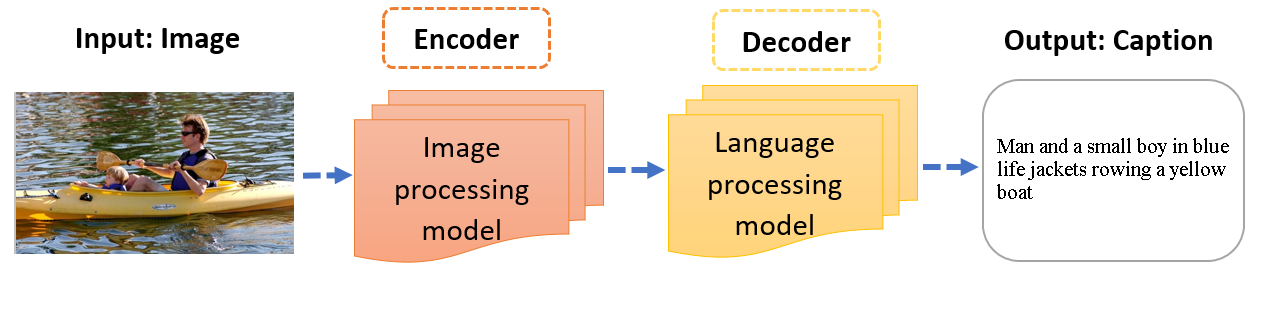}
    \caption{General architecture of image captioning model.}
    \label{fig:1st}
\end{figure*}

Identifying key components in an image, understanding their relationships, and creating syntactically and semantically consistent descriptions of the visual content are all necessary to create an image caption. This is one of the hardest tasks in artificial intelligence because it requires the integration of two very different research communities: natural language processing and computer vision \citep{hossain2019comprehensive}. An overview of the standard architecture of the image captioning model is given in Fig.\ref{fig:1st}. The general architecture of an image captioning system typically consists of several key components. It begins with an image input processed by a Convolutional Neural Network (CNN) for feature extraction, utilizing pre-trained models like ResNet or Inception to capture the essential visual elements. These extracted features are fed into a generation caption model such as Long-Short-Term Memory (LSTM) units or transformers. An optional attention mechanism can enhance this process by allowing the model to focus on specific image areas while forming each caption word. Finally, the system produces an output caption that represents the generated description of the image. 

Recent developments in deep learning models, made possible by cutting-edge computational capabilities, have significantly advanced this discipline \citep{cheikh2020active,yu2019multimodal}. Image captioning is challenging in artificial intelligence since it combines computer vision and natural language processing research. A captioning model aims to represent the text and scene, as this is essentially what the human brain does. Humans can automatically describe much information about any given image with a glance. One of the many difficulties and unsolved problems inherent in image captioning is the parallax error.
It may be difficult for the human eye to identify an object, even at certain angles, where its appearance varies to the point of being undetected. An object class may include several objects of various forms and angles. Additionally, the visual assistant could have difficulty correctly identifying objects hidden by other objects. Object recognition is negatively affected by scene clutter \cite{10.1145/3617592}. There are many industries in which image captioning research can find practical applications. Examples include medical imaging for analysis and diagnostics \citep{ayesha2021automatic, ogura2018effectiveness, depeursinge2017biomedical, al2010tumour}, improving education for students \citep{chendake2021learning, ayyoub2024learning}, supporting visually impaired people \citep{ahsan2021multi}, helping virtual assistants \citep{nivedita2021image}, facilitating information retrieval \citep{wang2020paic}, aiding video surveillance \citep{SALEEM2019108}, improving social media content \citep{shuster2019engaging}, and even assisting automated self-driving cars \citep{fujiyoshi2019deep}. Additionally, it is essential to improve the quality of image search \citep{guinness2018caption}. Template-based, retrieval-based, and deep learning-based approaches are the three primary categories of image captioning techniques. Template-based approaches create captions using predefined templates with blank spaces; this results in grammatically correct statements but is restricted. With retrieval-based techniques, general but sometimes inaccurate semantic descriptions are generated by extracting captions from an existing set. Using deep neural networks for visual and linguistic modeling in deep learning-based approaches is a discovery that improves image captioning systems and offers useful solutions. \citep{XIAO2024109626} introduced an attribute node to provide a more detailed description of objects and to model high-level relationships within a visual semantic graph. The proposed method of \citep{ZHANG2022108429} offers a novel approach to news image captioning, aiming to preserve semantic information, enhance style coherence with the news articles, and enable entity-aware, controllable caption generation. Before deep learning became popular, traditional machine learning approaches handled most image captioning tasks \cite{hossain2019comprehensive,bai2018survey}.
Among these were feature extraction approaches such as the histogram of Oriented Gradients (HOG), Local Binary Patterns (LBP), and Scale-Invariant Feature Transform (SIFT). The items were classified using a classifier after extracting features \cite{al2008combined,al2011supervised}. Deep learning-based techniques automatically find features and are more popular than traditional methods since feature extraction from huge amounts of data is challenging \cite{10.1145/3617592}. The use of deep machine learning for captioning images has received a lot of interest recently \cite{hossain2019comprehensive}. Deep learning algorithms can efficiently manage the difficulties and complexity of captioning images.
%Traditional learning methods often struggle to fully understand the intricate features and deep structure of data, which is critical to building effective models in data mining. 
Ensemble learning is a growing field of interest that addresses this issue by merging concepts from data fusion, modeling, and mining into a unified approach. It starts by extracting features using multiple algorithms that make predictions based on these characteristics. The ensemble learning then combines these insights to improve the general accuracy of the prediction through various voting mechanisms to achieve better results than any algorithm could provide \cite{dong2020survey}. Ensemble learning reduces the biases associated with individual models and enhances caption production overall by combining predictions from several models. This technique is particularly useful for improving the performance of complex architectures with several types of ensemble models, such as bagging, boosting, stacking, and voting.

The main contributions of this research are: (a) exploring the detailed design of transformer models, focusing on the effectiveness of different attention mechanisms,(b) introducing an innovative ensemble learning framework that leverages multiple deep neural network architectures to enhance the accuracy and richness of generated captions, and (c) enhancing the robustness and reducing overfitting of image captioning models by experimenting with different publicly available datasets.

%%%%%%%%%%%%%%%%%%%%%%%%%%%%%%%%%%%%%%%%%%%%%%%%%

The research is organized as follows. Section 2 describes related works on image captioning. Section 3 details the research methodology. Section 4 covers experiments and results. Section 5 provides limitations and future opportunities. Finally, Section 6 offers the conclusion and implications of the research.

%\section{Background}

\section {Related works}

Several papers have recently employed deep-learning techniques to generate captions for images. This section provides an overview and discussion of related works.

\subsection{Neural Network based model}
In the automatic generation of image captions\cite{verma2024automatic}, %the authors suggested a method that automatically creates captions for images using deep learning algorithms. 
the encoder-decoder architecture was used in the suggested model. The encoder processed the input image to extract the relevant data, while the decoder used these features to generate the caption.  %This CNN is pre-trained and optimized for localization and scene identification. 
The image was encoded into a feature vector with a specified length. Long Short-Term Memory (LSTM) cells were used to implement the decoder.% Flickr8k and MS-COCO Captions are two well-known captioning datasets that are used to train the model. These files include pictures with handwritten captions. The work applied accepted metrics to assess their model.
Utilizing pre-trained models and deep learning approaches, the suggested method showed encouraging results.% Researchers and practitioners interested in image captioning should investigate this work further to improve their models and applications.
The work in \cite{dahriimage} %presented a system that integrates computer vision and natural language processing to generate descriptive captions for images using neural network models. %The system used a CNN that has already been trained to extract pertinent information from the input images, these features represented the visual content. During the caption-generating process, some areas of the image were the focus of an attention-based method, this enables the model to generate descriptive sentences while focusing on the pertinent areas of the image. 
%Using the retrieved attention data and features, the RNN component created captions. %RNNs are appropriate for tasks that involve the creation of natural language because they perform well with sequential input. 
The suggested model performed better when creating informative captions for images. 
%Deep visual-semantic alignments
Authors of \cite{karpathy2015deep} provided a model that creates natural language descriptions of images for generating image descriptions. The method consisted of bidirectional RNNs over phrases, CNNs over image areas, and a structured objective using multimodal embedding to align the two modalities. %The work also presented a multimodal recurring neural network design that creates new descriptions of image regions based on detected alignments. 
The alignment model obtained state-of-the-art results. % in retrieval trials on Flickr8K, Flickr30K, and MSCOCO.
%Moreover, the generated descriptions of whole images and a new dataset with region-level annotations performed much better than the retrieval baselines.

%\subsection{Semantic based model}
%When creating image captions, authors of \cite{bineeshia2021image} addressed the difficulty of bridging the gap between semantic notions and visual aspects. They suggested a captioning technique that combines semantic and visual features. Convolutional Neural Networks (CNNs) were employed by the authors to extract visual information from input images. They also created an object recognition model in order to find semantic tags in the images. The encoder-decoder architecture was used in the suggested model. It used Long Short-Term Memory (LSTM) networks to generate captions after encoding both the visual features and the semantic data. Standard datasets like Flickr8k, Flickr30k, and MSCOCO were used to assess the model. Metrics like as BLEU and METEOR were employed to evaluate the quality of generated captions; their methodology incorporated visual and semantic data, resulting in enhanced mage captioning performance.\\

\subsection{Attention based model}
The challenging task of automatically producing meaningful captions for images was discussed in \cite{chu2020automatic}, and suggested a collaborative model known as AICRL (Automatic Image Captioning based on ResNet50 and LSTM with Soft Attention). The encoder used a CNN called ResNet50 to represent the input image comprehensively. %To generate captions later, ResNet50 first collected visual elements from the image and embedded them in a fixed-length vector. Long-Short-Term Memory (LSTM), an RNN architecture, was used in the decoder's construction. The decoder also included a gentle attention function. The model can predict the following sentence while focusing on particular areas of the image due to the soft attention technique. 
The model gathered the quality of generated captions by focusing on relevant locations. %Large datasets, such as the MS COCO 2014 dataset, which includes various images linked to descriptions created by humans, were used to train AICRL. 
The experiment results showed that the AICRL model is useful for producing image captions. It offers a promising means of bridging the gap between natural language descriptions and visual content, making it applicable to various computer vision applications and beyond.
It is remarkable that the aligned attention method is model-independent and may be quickly added to current innovative image captioning models to enhance their captioning capabilities. \cite{fei2022attention} presented a transformer-based model for image captioning; their strategy used a mask operation to automatically assess the influence of image region features and use the results as supervised information to direct attention alignment. %The basic version transformer was used, which performs N = 6 attention layers and employs h = 8 parallel attention heads each time. The researcher of this work investigated the link between attention weights and feature importance metrics in image captioning to provide a more thorough analysis regarding whether present attention mechanisms can focus on critical and effective image regions. The proposed model was evaluated using the Transformer model on the MS COCO dataset as a baseline. 
This work provided a useful reference for self-supervised learning. 
The transformer-based framework LATGeO was proposed in \cite{dubey2023label} to caption images, and it includes multi-level geometrically coherent and visual recommendations to relate objects based on their localized ratios. LATGeO used object proposals to find coherence and connected its embeddings with less significant surrounds. A brand-new label-attention module (LAM), an extension of the traditional transformer, was developed to bridge the gap between the visual and linguistic worlds. %In the suggested paradigm, each decoder layer's input includes object labels as background data for caption construction. Using just one model, comprehensive testing on the MS COCO dataset revealed improved results compared to alternative attention models. 
Although normalization has traditionally only been used outside of self-attention, the work of \cite{guo2020normalized} provided a unique normalization method and showed that doing so in hidden activation within self-attention is feasible and advantageous. They provide a class of geometry-aware self-attention (GSA) that extends self-attention to explicitly and efficiently consider the relative geometry relations between the objects in the image to model the geometry structure of the input objects for feature extraction. Faster-RCNN was used. The inputs to the transformer encoder are region-based visuals, and the transformer decoder predicts the subsequent word recursively using the attended senses and the embedding of the preceding words. %The generality of this method is demonstrated by experiments using the MS-COCO dataset on three difficult tasks: video captioning, machine translation, and visual question answering. 

Motivated by the relationships between image features, \cite{he2020image} presented a new transformer-based model. The proposed model considered three types of spatial relationships in the image regions. The query region could be a parent, neighbor, or child. %Spatially adjacent matrices were used to combine the output of the parallel sub-transformer layers. 
The decoder consists of an LSTM layer and an implicit transformer layer. The transformer was used parallel to decode different image regions in the decoder part. %The model was tested on the MSCOCO online dataset with 40,775 images and, compared to current methods. 
The results showed that the proposed model was better than others based on several evaluation metrics.% Qualitative and quantitative analysis were provided to validate the model. This model can help researchers with computer vision tasks. 
The work of \cite{yu2019multimodal} illustrated the limitations of current methods, such as neglecting the interaction between a word and an object and the undiscovered relationship between objects. To solve these problems, they presented a multi-transformer (MT) for image captioning. The MT model can understand three types of relations: word-to-word, object-to-object, and word-to-object. %The transformer mechanism consists of an image encoder and a text decoder. The image encoder contains two parts: an aligned multiview encoder (AMV) and an aligned multiview decoder (UMV). The AMV model used a pre-trained R-CNN to extract region features; each pair of features was aligned to one object in the image. To improve the AMV model, they added a UMV encoder model that can integrate features from different object detectors. 
The caption decoder took the encoder output and generated the caption using word embedding and a layer of LSTM. %The proposed model was tested on the MSCOCO 2015 dataset. The results showed an improvement in the existing approaches.

%%%%%%%%%%%%%%%%%%%%%%%%%%%
\subsection{Ensemble based model }
Ensemble learning aims to increase generalizability and robustness over a single model by combining the predictions of various base models. Modern techniques for detecting hate speech in multimodal memes \cite{ velioglu2020detecting} applied the majority voting technique, also known as the hard voting or voting classifier, which combines many classifiers and voting classifiers. %Class labels can be predicted using a majority vote to counteract the limitations of many models that perform equally well, and the resulting classifier is frequently helpful. 
As a result, it performs better than any individual model utilized in the ensemble. 
Textual and visual hybrid methods are combined using the max voting technique to classify a fake or real news instance. \cite{ meel2021han}, in this work, the maximum voting method was used. The proposed system consists of four independent parallel streams capable of detecting specific forgeries. All four streams handled each input instance. These independent predictions are finally combined using the maximum voting ensemble method. %This method classifies a news instance as true if more than half of the base techniques predict it as true; otherwise, it is false. If two predictions are real and two are fake, the overall prediction is fake. The ensemble design has the benefit of significantly increasing the accuracy of the hybrid multimodal false news detection framework.\\

In \cite{zhong2022multi}, an image captioning method was presented using a set of weighted multichannel fusion optimization enhancements to optimize the encoder and decoder. In the model that is being described, a multichannel encoder was suggested that can combine different models and algorithms to extract different information from the same image, %To explain how the decoder receives features from several channels, 
researchers suggested combining separate decoders of the same type using the voting weight technique for decoder fusion to improve the description produced by the decoder. 
For the concept detection task, \cite{dalla2022cmre} considered an image retrieval approach using an ensemble of five different CNNs, where the top $N$ photos most similar to the training set and their related CUIs were used to assign a set of CUIs to each query image. The top $N$ images that look the most like a query image, determined by the cosine similarity between image embeddings, were extracted using CNN as the image encoder; then, an aggregation step was carried out to choose the set of CUIs to link to each query image. This involved soft majority voting.
%, in which a CUI is given to the query image if it occurs in at least 30\% of the returned images. These stages were completed individually for the DenseNet-201 models. Lastly, a set of five DenseNet-201 models is used to allocate the union of the anticipated CUIs from each model to each image. 
A recent work \cite{ salur2022soft} proposes a soft voting-based ensemble model that benefits from the efficient operation of various classifiers on various modalities. Deep feature extraction from multimodal datasets was performed for the proposed model using deep learning methods (BiLSTM, CNN). The final feature sets were classified using the soft voting-based ensemble learning model after completing the feature selection process for the features that combine text and image features.% The experimental experiments showed that the suggested model outperformed other adversarial models. 
In \cite{singh2022efficient}, an effective deep-set medical image captioning network (DCNet) was suggested to give doctors and patients explanations. Three well-known pre-trained models, including VGG16, ResNet152V2, and DenseNet201, are combined into DCNet. Assembling these models leads to better outcomes, as it avoids an overfitting problem. 
A classifier was created according to the research conducted by \cite{kim2022inspection} using a soft voting ensemble combining the common CNN models %DenseNets, EfficientNets, Inceptions, MobileNets, ResNets, and VGGs. In the soft voting ensemble classifier, various models were combined. These models combine individual decisions based on probability values to designate that the data belong to a given class.
Predictions in the soft voting ensemble are combined and weighted according to the relevance of the classifier to produce the total of weighted probabilities.
%%%%%%%%%%%%%%%%%%%%%%%%

\subsection{Insights from previous research and our solution}
%%%%%%%%%%%%% just highlighted and have a new section
The discussion above revealed that contemporary captioning models rely on RNN and LSTM as language models. However, one key issue with these approaches is the occurrence of vanishing gradients, limiting their effectiveness. Moreover, the RNN and LSTM models are not hardware-friendly and require additional computational resources. An alternative approach explored in the literature is using Generative Adversarial Networks (GAN) for image captioning. However, GANs come with challenges due to their discrete nature, making training such systems a difficult task \cite{abu2022effect,abu2021paired}. Using a hybrid approach, combining LSTM with transformer models introduces specific limitations and drawbacks. For example, it can increase the complexity of the model, attributed to architectural differences, resulting in a higher demand for resources and extended training times \cite{alkadi2020deep}. Consequently, this complexity can affect the interpretation of model decisions, hindering a clear understanding of the underlying reasoning.
Image captioning is an attractive task that involves understanding visual and textual information. The need for image captioning arises from the need to make visual content accessible to individuals. Therefore, developing and implementing dedicated image captioning systems is essential to address this need. Therefore, this research aims to bridge this gap by introducing a hybrid approach that combines a transformer with an attention mechanism to help the model capture complex details in images and generate more contextually relevant captions. The rationale behind this combination is that transformers are great for capturing long-range dependencies in data, while attention mechanisms help them focus on relevant parts. Ensemble learning, on the other hand, can boost overall performance by combining multiple models. The subsequent section will explore the details of this approach.

\section{Methodology}
\label{Meth}
This work followed a methodology incorporating four stages: data description and data preprocessing, model development, experimentation, and performance evaluation. The following subsections discuss each stage in more detail.

\subsection{Dataset}

This section will introduce the commonly used datasets in image captioning. %Table \ref{tab:dataset} shows the
details of these datasets.

%\begin{longtable}[c]{ccccc}
%%\caption{Publicly used datasets for image captioning}
%\label{tab:dataset}\\
%\hline
%\textbf{Datasets}  & \textbf{Train} & \textbf{Validate} & \textbf{Test} & \textbf{Captions} \\ \hline
%\endfirsthead
%
%\multicolumn{5}{c}%
%{{\bfseries Table \thetable\ continued from previous page}} \\
%\hline
%\textbf{Datasets}  & \textbf{Train} & \textbf{Validate} & \textbf{Test} & \textbf{Captions} \\ \hline
%\endhead
%
%\textbf{Flickr8k}  & 6,000          & 1,000             & 1,000         & 5                 \\ \hline
%\textbf{Flickr30k} & 29,783         & 1,000             & 1,000         & 5                 \\ \hline
%\textbf{MS COCO}   & 113,287        & 5,000             & 5,000         & 5                 \\ \hline
%\end{longtable}

%\begin{figure*}[h]
%\centering
%\includegraphics[width=\textwidth,height=\textheight,keepaspectratio]{Flickr8k.png}
%\caption{Exemplary instances extracted from Flickr8k dataset}
%\label{fig:Flickr8k}
%\end{figure*}

\textbf{Flickr8K}\cite{hodosh2013framing}: it was published for public use in 2013. The photographs in the dataset, which total 8000, are all from the photo and image-sharing website Flickr. The image content is mostly human and animal. The description for the label was also crowd-sourced through Amazon's manual labeling program. Each image contains a description of five sentences. This dataset offered a comprehensive and diverse set of images comprising 6,000 training images, 1,000 validation images, and 1,000 test images. Flickr8k is a standard dataset for training and evaluating image captioning models, covering a wide range of scenes, objects, and activities characteristic of daily photography. Researchers use its rich diversity in images and textual descriptions to develop algorithms capable of generating accurate and contextually relevant captions.

\textbf{Flickr30k}\cite{young2014image}: Flickr8k dataset has been expanded to build Flickr30k, it contains 31,783 captioned images. The split dataset available to the public uses 29,000, 1,000, and 1,000 images for training, validation, and testing, respectively. Each image has five sentences that were written specifically for it. The photos in this dataset mostly show people participating in ordinary activities and events. Flickr30k is used to understand visual media (images) that match a language expression (an image description). This dataset is frequently used as a reference standard for sentence-based image descriptions.  A research paper emphasizes the importance of the Flickr30K dataset in analyzing human descriptions of visual content, providing a comprehensive review of its features. Each image is richly annotated with contextually relevant descriptions that offer multiple viewpoints on its content. The dataset captures the diversity of human experience and includes various aspects of human actions, objects, scenes, and environments. This variety makes it particularly suitable for exploring how people interpret and describe visual scenes \cite{38c45b74b30942f89412e6008ad3db1b}.

%Fig.\ref{fig:Flickr30k} shows example of instances extracted from Flickr8k dataset.

%\begin{figure*}[h]
%\centering
%\includegraphics[width=\textwidth,height=\textheight,keepaspectratio]{Flickr30k.png}
%\caption{Exemplary instances extracted from Flickr30k dataset}
%\label{fig:Flickr30k}
%\end{figure*}

%%%%\subsection{Microsoft COCO The MS COCO }Microsoft's COCO dataset \cite{lin2014microsoft} is a large-scale dataset that may be utilized for applications including image recognition, object detection, semantic segmentation, and image captioning. Each image in the dataset is artificially tagged using Amazon Mechanical Turk, and it contains over 100 object categories derived from images of complex daily situations containing common things against natural backgrounds (AMT).The dataset has 82,783 training image samples and 40,504 verification image samples. Furthermore, there are 40,775 test photos with labels that are not available to the public. There are five caption sentences for each image. \\%%%

% \usepackage{longtable}
% Note: It may be necessary to compile the document several times to get a multi-page table to line up properly

\subsection{Data preprocessing}
\label{4.2}
Because of raw textual data challenges, cleaning and preprocessing datasets before they are used in ML models have become essential. The approach we applied to text preprocessing was comprehensive and systematic. Several procedures were employed in the data preprocessing, including the following:

\begin{enumerate}[label=(\alph*)]
    \item Text normalization: Typically, actions are taken to reduce the number of extracted terms. They include eliminating special and non-letter characters (\$, \&, \%,.. ).

    \item Text tokenization: In this step, a linguistic analysis of the text is performed. Separates words, character strings, and punctuation marks into tokens during indexing. This process aims to divide the text into a stream of discrete tokens, or words, by identifying the sentences' borders and eliminating any unnecessary punctuation.

    \item Adding start and end tokens: Finally, distinctive start and end tokens were appended to determine the beginning and end of each caption, adding a layer of structural clarity to the dataset. A unique padding token was introduced to address the variability and standardize the length of captions.
\end{enumerate}

\subsection{ The proposed model for image captioning}
%Fig. \ref{fig:framework} shows the schematic diagram of the proposed model.

The following are the steps applied through the model, and the following subsections discuss each stage in more detail. Algorithm \ref{alg:ARIC} provides the pseudo-code outlining the operations of the model.

\begin{enumerate} [label=S\arabic*:] 
    \item Image feature extraction: A pre-trained CNN network like ResNet or EfficientNet extracts features from the input image. These features serve as a rich representation of the visual content.
    \item Text generation with a transformer: A transformer-based model generates textual descriptions by taking as input the image features and producing a sequence of words that form the caption.
    \item Attention Mechanism: Attention mechanisms are implemented within the transformer, allowing the model to focus on different parts of the image when generating each word in the caption. It enhances the model's ability to align visual and textual information.
    \item The Beam Search Algorithm: The beam search algorithm was applied with a width of $k$ = 10.
    \item Ensemble learning: To get a more robust and accurate caption, the ensemble learning model trains multiple instances of the transformer with different random initializations or hyperparameters and then combines their output, either by averaging or voting.
    \item Training and fine-tuning: Train the combined model on a large dataset of image-caption pairs, then fine-tune the model on a specific dataset.
    \item Evaluation: Evaluate the performance of the ensemble model using metrics like BLEU, METEOR, and CIDEr.
\end{enumerate}

\begin{algorithm*}[ht]
   \caption{Attention-based transformer model using ensemble learning}
   \label{alg:ARIC}
    \begin{algorithmic} [1]
    \INPUT { dataset=[Set of images ($S$), corresponding set of captions ($CI$)]}
    \OUTPUT  {The final output caption for the tested image ($o_c$)}
    \STATE {Evaluation metrics: (BLEU-[1-4], ROUGE-L, METEOR, CIDEr, \& SPICE)}
    \STATE \textbf{Step1: Dataset prepossessing} 
       \FOR{each caption set $CI$ of an image $I$}  
         \STATE Normalization ($CI$)
         \STATE Text tokenization ($CI$)
         \STATE Adding start and end tokens ($CI$) <start> ($CI$) <end>
       \ENDFOR
      \FOR{each image $I \in S$}
        \STATE Augment ($I$)
       \ENDFOR
    \STATE \textbf{Step2: Feature extraction}
      \STATE $M$= [ResNet50, ResNet101, EfficientNetV2, VGG16, VGG19, EfficientNetB4, ResNet152, RegNetX120]
      \FOR{each image $I \in S$}
        \FOR{each pre-trained model $m \in M$}
          \STATE $f_i$ = extract feature map $f_i$ of image $I$ 
        \ENDFOR
      \ENDFOR
    \STATE \textbf{Step3: Caption generation} 
      \FOR{each feature map $f_i$ }
        \STATE $g_c$= generatedCaptionbyTransformer
        \STATE BestKCaption= Beam Search(10)
     \ENDFOR
    \STATE \textbf {Step4: Ensemble learning} 
     \FOR {each $g_c$}
        \STATE $o_c$= voting-on (the generated caption from all models $g_c$)
     \ENDFOR
 \end{algorithmic}
\end{algorithm*}

%Contemporary image captioning models have largely incorporated a flexible and effective encoder-decoder architecture, often called a CNN+RNN structure. The architecture of the proposed model consists of two primary models: the image-processing model and the language-processing model. In this configuration, the encoder typically employs a CNN image model to extract high-level feature vectors from input images and effectively ``reads'' them. Meanwhile, the decoder, often implemented as RNN, generates words based on the image representation acquired from the encoder. Its task is to produce a sequence of words that form a coherent, grammatically correct, and stylistically accurate phrase, effectively encapsulating the image's content \cite{staniute2019systematic}.

%The proposed model in this study adopts an encoder-decoder approach enriched with attention mechanisms, as \cite{xu2015show} recommended. The attention mechanism focuses on relevant sections of the image vital for the caption creation process, potentially leading to superior outcomes. Fig. \ref{fig:1st} provides an overview of the typical architecture of image captioning model.

\subsubsection{Image feature extraction}

\begin{figure*}[ht]
\centering
\includegraphics[width=7cm]{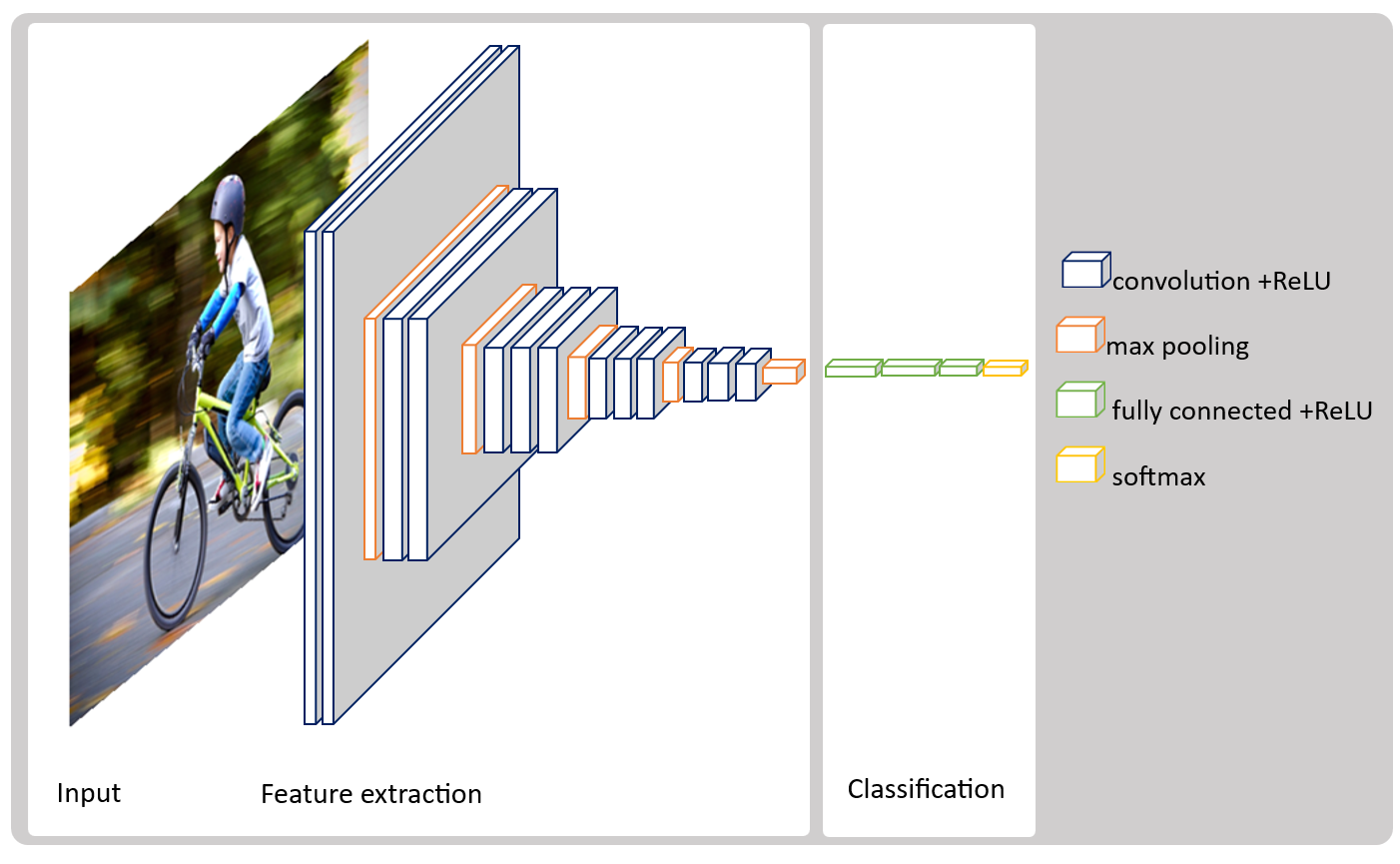}
\caption{ Architecture of CNN Model }
\label{cnn}
\end{figure*}

%%%%%%%%%%%%%%%%
%\subsection{Attention in image processing}
\textbf{a) Convolutional neural network (CNN):} Popular deep learning models include recurrent neural networks (RNNs), convolutional neural networks (CNNs), deep belief networks (DBNs), and deep Boltzmann machines (DBMs). Using shared weight filters and hierarchical learning, CNNs are highly effective in understanding visual data \cite{oluwasammi2021features,mnr2021}. CNN-based encoders on ImageNet that have been pre-trained are frequently used in image captioning to convert images into visual vectors. Selective focus during generation is made possible by preserving fine-grained correspondence using sets from lower convolution layers \cite{ biswas2020towards,chen2022news}. 
CNNs provide the following benefits over conventional neural networks when used in computer vision applications: 1) The main reason to consider CNN is its weight-sharing feature, which reduces the number of trainable network parameters, allowing the network to increase generalization and preventing overfitting. 2) Learning both the classification layer and the feature extraction layers simultaneously produces a well-structured model output that depends on the features that were extracted. 3) CNN facilitates large-scale network installation more easily than other neural networks \cite{alzubaidi2021review}. See Fig. \ref{cnn} that presents the architecture of the CNN model.

\noindent \textbf{b) Transfer learning:} Applying a previously learned model to a modified environment is known as transfer learning. Due to its ability to train deep neural networks on tiny datasets, it is particularly well preferred in the deep learning field. This is particularly helpful in data science because most real-world scenarios do not require millions of labeled data sets to train complicated models. To apply transfer learning to image captioning, the model was first trained on a standard dataset under supervision, and then its knowledge was transferred to a new dataset consisting of unpaired phrases and images \cite{oluwasammi2021features}.
%\indent 
Residual CNNs, such as ResNet-50, use identity mapping and shortcut connections to address overfitting and optimization issues. A pre-trained ResNet-50, trained on ImageNet, is utilized in image feature extraction by removing its final output layer \citep{faiyaz2021improved}. The ResNet-101 image captioning model uses bottom-up attention to encode images as a baseline. The effectiveness of bottom-up attention to the baseline ResNet encoding is evaluated to evaluate the performance of the model \citep{anderson2018bottom}. Shorter connections between layers in DenseNet's architecture improve training efficiency and the depth of deep learning networks. Strong information flow is ensured by interlayer connection, which improves learning \citep{huang2017densely}. However, VGGNet is a popular image feature extractor that is frequently used in research applications because of its resilience and simplicity. ResNet, however, outperforms VGG in terms of efficiency, providing better accuracy with fewer parameters \citep{staniute2019systematic}. Compound-scaling EfficientNet models have recently proven superior to other CNNs' accuracy and efficiency when used with transfer learning datasets. They show promise in various fields, such as the classification of COVID-19 \cite{tan2019efficientnet, marques2020automated}.
MobileNet is another architecture that maximizes computational effectiveness while maintaining good accuracy. The effectiveness of representation is improved by channel separation and reintegration \citep{sandler2018mobilenetv2, INDRASWARI2022198}. Lastly, Inception-v3, often used for transfer learning, has less computational overhead when used as an encoder. It gathers key information and adds it to a feature matrix that captures the essence of an image \cite{ yu2019multimodal,do2020reference, he2020image}. In the proposed methodology's workflow, the initial step toward image processing involves passing the image through a CNN to generate image features. Existing work studied various versions of CNN as feature extractors for image captioning. Feature extraction is based on eight CNN models discussed in Section. These models include: ResNet50, ResNet101, EfficientNetV2, VGG16, VGG19, EfficientNetB4, ResNet152, and RegNetX120. These features serve as input for the subsequent language processing model. 
%%%%%%%%%%%%% just illustration was highlighted to be clear  Fig.4:  R1.7
Fig. \ref{cnn} visually depicts the CNN model architecture. The convolution layer plays a vital role in downsampling the image into features and incorporating information from nearby pixels. The prediction layers then become active, using multiple convolution filters or kernels that pass over the image, each extracting unique aspects. To prevent overfitting and reduce the spatial size of the convolved features, a max pooling layer is used to provide an abstract representation of the convolved features. ReLU is the most widely used among various activation functions due to its ease of training and superior performance attributed to its linear behavior, as highlighted by \cite{alzubaidi2021review}.

\subsubsection{Text generation with a transformer} 

%%%%%%%%%%%%%%%%%%%%%%%%%%%%%
%\subsection{Transformers}

One kind of neural network architecture is a transformer. Transformer was first introduced in the publication ``Attention is all you need'' \cite{vaswani2017attention}. Text data is well handled by the Transformer architecture, which is sequential by design. After receiving one text sequence as input, they create another one with a stack of encoder and decoder layers. The encoder and decoder stacks contain matching embedding layers for their respective inputs. There is an output layer at the end to create the final result. The encoder and a feedforward layer contain the crucial self-attention layer, which determines the connections between the words in the sequence. The decoder consists of the feedforward layer, the self-attention layer, and a second encoder-decoder attention layer. There is a distinct set of weights for each encoder and decoder. Current image captioning algorithms get an excellent score by intuitively connecting informative parts of the image with transformer designs and attention. Some earlier transformer-based image captioning models, however, are limited in their ability to use the basic machine translation architecture of the transformer. A word in a text can be located to the left or right of another word, depending on how far apart they are. The degree of freedom in the relative spatial relationship between areas in images is more significant than in phrases \cite{he2020image}. This is because images are two- or three-dimensional, meaning a region might be anywhere besides the left or right of another region.

\begin{figure*}[ht]
\centering
\includegraphics[width=8cm]{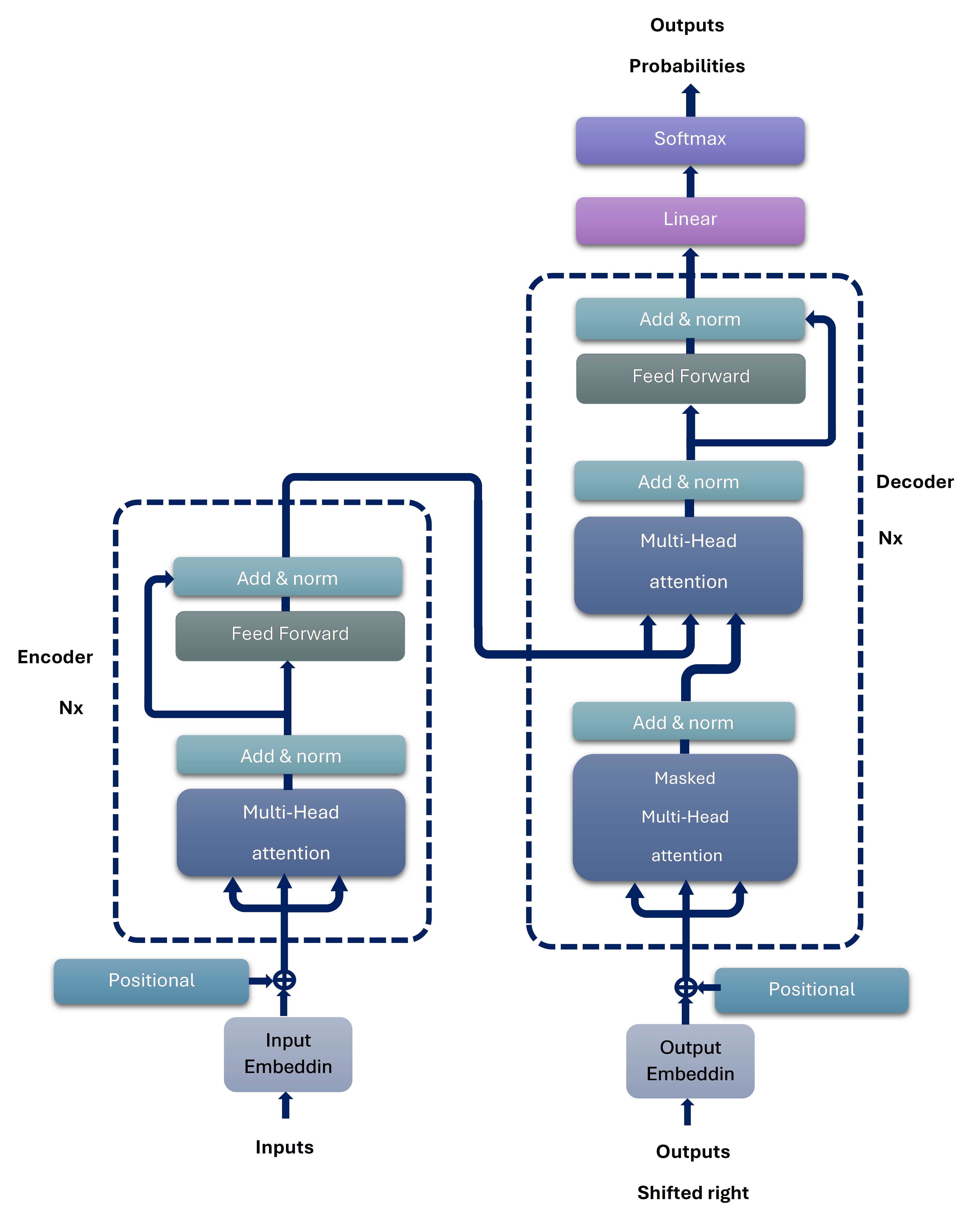}
\caption{General Architecture of Transformer }
\label{Transformer_2}
\end{figure*}

%%%%%%%%%  illustration weas highlighted to be clear Fig2 and Fig.3 R1.7
An encoder and a decoder are the two primary components of the transformer, as depicted in Fig. \ref{Transformer_2}. Similar to parallel heads of self-attention, multi-head attention functions. The transformer uses self-attention to incorporate its understanding of other relevant terms into the word it is currently processing. The fully connected feedforward network is an additional component that consists of two linear transformations with different parameters at different layers, but that is the same across positions. To help with word position determination, the transformer adds a vector to each input embedding. Position embedding is a technique that considers the order of words in an input sequence. The vector generated by the decoder's stack is transformed into a larger vector known as a logit vector by the linear layer, a straightforward, fully connected neural network. The probabilities are supplied by SoftMax. The term related to the cell of highest probability is generated as output \cite{vaswani2017attention}.

\begin{figure*}[ht]
\centering
\includegraphics[width=9cm]{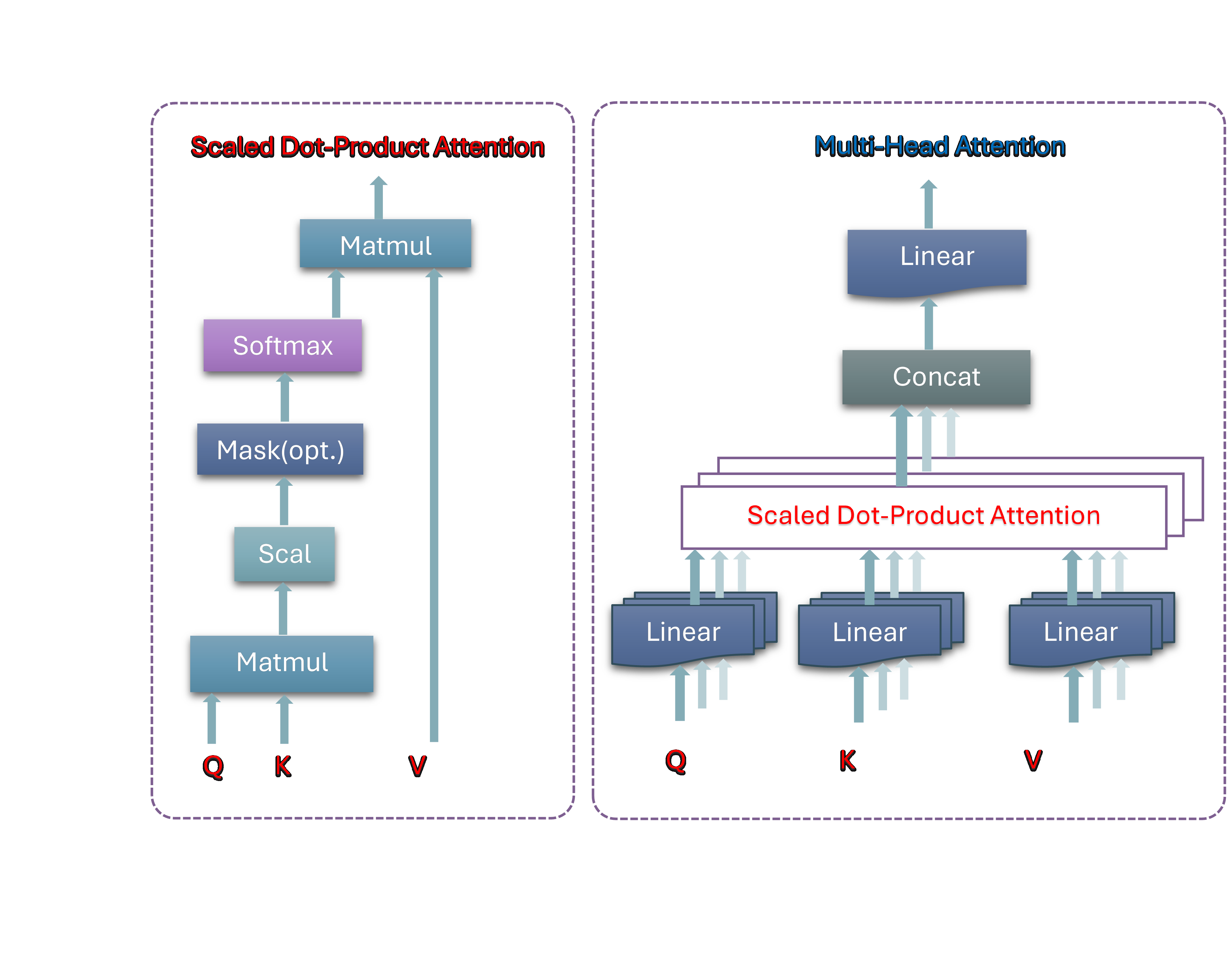}
\caption{Scaled Dot Product Attention (left),  Multi-Head Attention (right)}
\label{scaled-multi}
\end{figure*}

To obtain the attention scores, Fig.\ref{scaled-multi} shows the scaled attention of the dot product in the left block, in which the self-attention computes the dot product of the query with all keys, which is then normalized using the SoftMax operator. The attention scores determine the weights, and each entity then becomes the weighted sum of all the entities in the sequence. On the other hand, on the right block, multihead attention consists of numerous self-attention blocks (h = 8 in the original Transformer model) to capture multiple complex interactions between various items in the sequence.

%%%%%%%%%%%%%%%%%%%%%%%%%%%%

The language processing model encompasses three components: the transformer, the attention mechanism, and the ensemble learning model.  A transformer-based model generates textual descriptions by taking the image features as input and producing a sequence of words that form the caption. In this work, the proposed language processing model uses the transformer with two key components: the encoder and the decoder (refer to Fig. \ref{Transformer_2}). The image transformer utilized for image captioning will decode diverse information within image regions \cite{gong2019enhanced}. To establish the position of each word, the transformer introduces a vector added to each input embedding. Position embedding accounts for the sequential order of words in an input sequence. The linear layer, a straightforward, fully connected neural network, transforms the vector generated by the stack of decoders into a substantially larger vector referred to as a logit vector. Subsequently, SoftMax is applied to derive probabilities. The cell with the highest probability is selected, and the associated word becomes the output \cite{vaswani2017attention}. The transformer model addresses issues inherent in RNN and LSTM, facilitating increased parallelization and enhancing translation quality. Unlike LSTMs or RNNs, which process sentences one word at a time, transformer models are attention-based, capable of handling entire sentences \cite{wolf2020transformers}.

\subsubsection{The attention mechanism}

%%%%%%%%%%%%%%%%%%%%%%%%%%%%%
%\noindent\textbf{c) Attention mechanisms:}  
Attention mechanisms focus on the most relevant features extracted by CNNs, which is crucial for tasks such as image captioning, where context is key. Attention in image processing mimics human attention patterns. Its strength lies in establishing meaningful connections between features and enhancing the models' ability to prioritize important features while filtering out noise. This aligns with the attention mechanisms that guide the focus of the model during training \cite{10.1145/3617592}. Despite the richness of the image data, not all features require explicit attention in captioning. When attention is integrated into the encoder-decoder picture captioning framework, sentence creation becomes contingent on hidden states computed using the attention method. The attention mechanism is a fundamental component of the encoder-decoder architecture within this framework. Using various types of input image patterns to guide the decoding process, ensuring that attention is focused on specific features of the input image at each time step. This composed focus on attention facilitates the generation of a descriptive caption for the input image \cite{BAI2018291}.

Attention guides computations on significant regions to improve caption quality in image annotation. This is achieved by using soft and hard attention mechanisms to estimate the focus of attention. Soft attention, trainable via standard backpropagation, involves weighting the annotated vector of picture features when salient features are identified. On the other hand, stochastic hard attention is trained by maximizing a variation lower limit \cite{oluwasammi2021features}. Recent studies have explored top-down and bottom-up attention theories, with recent experiments favoring top-down attention mechanisms \cite{staniute2019systematic}. Attentive encoder-decoder models lack global modeling skills. To address this, a reviewer module reviews encoder hidden states, producing a thought vector at each step. The attention mechanism plays a vital role in assigning weights to hidden states. These thought vectors capture global input aspects and effectively review and learn the encoded information from the encoder. Subsequently, the decoder uses these thought vectors to predict the next word in the sequence \cite{BAI2018291}. Visual attention in multimodal coverage mechanisms bridges the gap between encoder and decoder, improving data understanding \cite{chen2019news, biswas2020towards}. Scaled \textit{Dot Product Attention}, introduced by \cite{vaswani2017attention}, computes the dot products of the queries, the dimensions keys $d_k$, and the dimensions $d_v$ values that make up the input; after that, the dot products of the query with all the calculated keys, divided by $\sqrt{d_k}$, and then a SoftMax function was applied to obtain the weights of the values. The attention function was continuously computed on a group of queries gathered into a matrix $Q$. The keys and values are also compacted in matrices $K$ and $V$. The attention function is mathematically formalized in (\ref{equ:scaleddotAtt}).

\begin{equation}
\label{equ:scaleddotAtt}
\operatorname{Attention }(Q, K, V)=\operatorname{SoftMax}\left(\frac{Q K^T}{\sqrt{d_k}}\right) V
\end{equation}

\noindent The mechanism of multi-head attention involves parallel passes through the attention mechanism. The formula expressed in (\ref{equ:MultiHAtt}) produces concatenated outputs to be transformed into the desired dimension \cite{vaswani2017attention}.

\begin{equation}
\label{equ:MultiHAtt}
\begin{aligned}
\operatorname{MultiHead}(Q, K, V) & =\operatorname{Concat}\left(\operatorname{head}_1, \ldots, \operatorname{head}_{\mathrm{h}}\right) W^O \\
\text {where head} & =\operatorname{Attention}\left(Q W_i^Q, K W_i^K, V W_i^V\right)
\end{aligned}
\end{equation}

%%%%%%%%%%%%%%%%%%%%%%%%%%%

\begin{figure*}[ht]
\centering
\includegraphics[width=0.7\textwidth,height=\textheight,keepaspectratio]{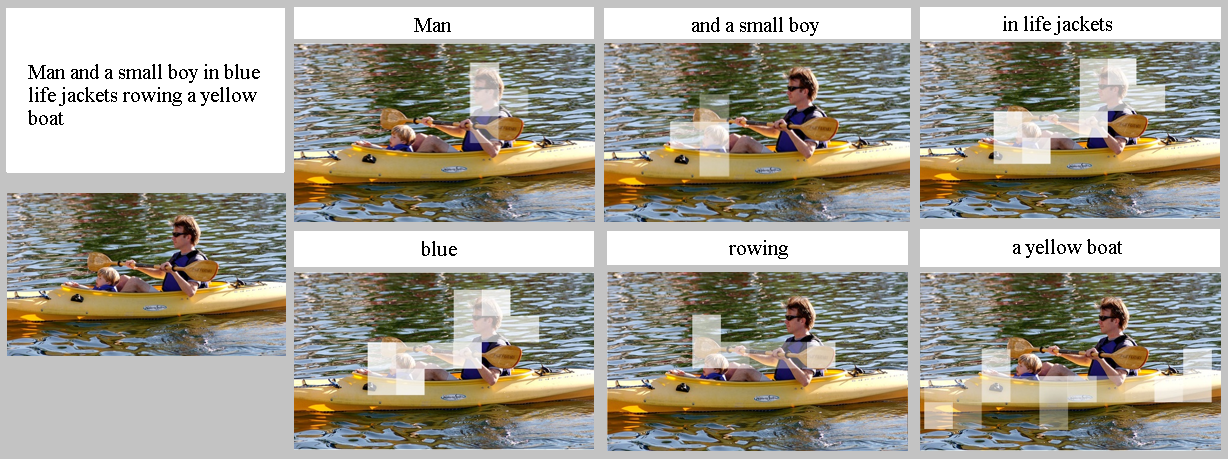}
\caption{The visual architecture of the convolutional neural network.}
\label{fig:attn}
\end{figure*}
The attention mechanism is implemented within the transformer, which allows the model to focus on different parts of the image when generating each word in the caption. It enhances the model's ability to align visual and textual information. Generally, individuals selectively attend to information, focusing on secondary data while disregarding certain primary data. This attention mechanism is essential for generation-based models within the encoder-decoder architecture, mirroring human visual focus in image captioning. In cognitive neurology, attention is identified as a shared higher cognitive skill that allows intentional oversight of received data. Originally proposed for image categorization, attention is widely used in NLP experiments, including machine translation, speech recognition, text understanding, and visual captioning \cite{pedersoli2017areas,MISHRA2021107114,yu2019multimodal,wang2020overview}.

Fig. \ref{fig:attn} visually depicts attention over time, illustrating how the model's focus shifts with the generation of each word to highlight relevant parts of the image. CNN highlights different areas of the image to draw attention to key components that are essential for correct interpretation. This detailed analysis demonstrates how CNNs analyze visual data to properly interpret and caption images. CNN starts by analyzing the image to find fundamental elements like borders and color spots. In this instance, it could pick up on the boy's and man's shapes and the boat's recognizable yellow color. More intricate characteristics are found; the network could identify the forms of a boat and a life jacket. As it gets farther into the network, CNN combines basic properties to detect things. The little child and the man are depicted as distinct individuals, each with a blue life jacket. The actions included in the image, through further analysis, acknowledge that the individuals are taking part in an activity (rowing). Combining all recognized details allowed for a comprehensive understanding of the scene about a ``man and a small boy in blue life jackets rowing a yellow boat''.

\subsubsection{The Beam search algorithm} 

The greedy decoding technique outputs the word with the highest probability. However, it quickly accumulates potential errors. To solve this problem, the beam search algorithm with a width of $k = 10$ was applied, maintaining $k$ sequence candidates and selecting the most likely one at each step \cite{stefanini2022show}. This approach generates a diverse group of captions. Previous studies supported beam search as the preferred algorithm for caption generation \cite{Li2016ASF}.

\subsubsection{Ensemble learning}

%%%%%%%%%%%%%%%%%%%%%%%%
%\subsection{Ensemble learning}

Typical learning techniques may not produce sufficient results because various features and the underlying structure of data are difficult for these methods to capture. So, building an effective model becomes a crucial problem in the data mining industry. An area of research that is gaining interest is ensemble learning, which tries to combine data fusion, data modeling, and data mining into a single framework. In which a set of features is first extracted using multiple learning algorithms to provide predictions based on these learned properties. Then, ensemble learning combines useful information to improve prediction accuracy across a variety of voting processes to outperform the results of any individual algorithm. Through the use of multiple machine learning algorithms, ensemble learning techniques generate weak predictions based on features extracted from a variety of data projections. The results are then fused with different voting mechanisms to produce performances that are better than those of any one of the constituent algorithms alone. Ensemble learning is used to improve architecture performance \cite{dong2020survey}. Several ensemble models exist\cite{mienye2022survey}, including bagging, boosting, stacking, voting:\\

\textbf{a) Bagging:} Breiman \cite{breiman1996bagging} created bootstrap aggregation, often known as bagging, to improve the classification performance of machine learning models by aggregating the predictions from randomly generated training sets. It was argued that bagging can increase accuracy because varying the learning set can result in appreciable changes to the predictor that is produced. In addition, diversity is achieved in bagging through the creation of bootstrapped copies of the input data, in which a number of randomly selected subsets are selected with replacements from the initial training set. As a result, the different training sets are considered distinct and are used to train different base learners for the same machine-learning algorithm.\\

\textbf{b) Boosting:} A machine learning method called ``boosting'' can turn a weak classifier into a powerful one. It is a kind of ensemble meta-algorithm that lowers variance and bias. A classifier that performs marginally better than random guessing is considered weak, whereas classifiers that achieve considerable accuracy are considered strong, and it is upon these classifiers that the boosting ensemble methods are based. \cite{schapire1990strength} addressed the boosting algorithm regarding the possibility of producing a single strong learner from a group of weak learners. \\

\textbf{c) Stacking:} An ensemble learning framework called stacked generalization, or stacking, trains a different machine learning algorithm to aggregate the predictions of two or more ensemble members. Wolpert \cite{WOLPERT1992241} first proposed an effort to reduce the generalization error in machine learning issues. When many machine learning models are particularly skilled at a specific position, stacking can be helpful. In this case, the stacking strategy uses a different ML model to determine when to employ the predictions from the different models. It entails training a meta-learning algorithm to train a new model that combines the predictions from the base models with numerous base algorithms, the so-called level-0 models.\\

%%%%%%% just highlighted subsection C1.2
\textbf{d) Voting:} In problems involving regression and classification, the majority vote is the most widely used and logical combination approach \cite{BALLABIO2019129}. The class with the majority vote is returned as the ensemble prediction in classification problems when the predictions for each class are added together. Meanwhile, regression tasks determine the majority vote by averaging the predictions made by each base learner. Let us assume that the $t$ classifier's decision is $d_{t, c} \in\{0,1\}$, $t = 1,\dots, T$ and $c = 1,\dots, C$, where $T$ and $C$ represent the number of classes and classifiers, respectively. Next, class $\omega_c *$  is chosen as the ensemble forecast by majority voting if

\begin{equation}
\sum_{t=1}^T d_{t, c}=\max _c \sum_{t=1}^T d_{t, c}
\end{equation}

\begin{figure*}[ht]
\centering
\includegraphics[width=0.42\textwidth,height=\textheight,keepaspectratio]{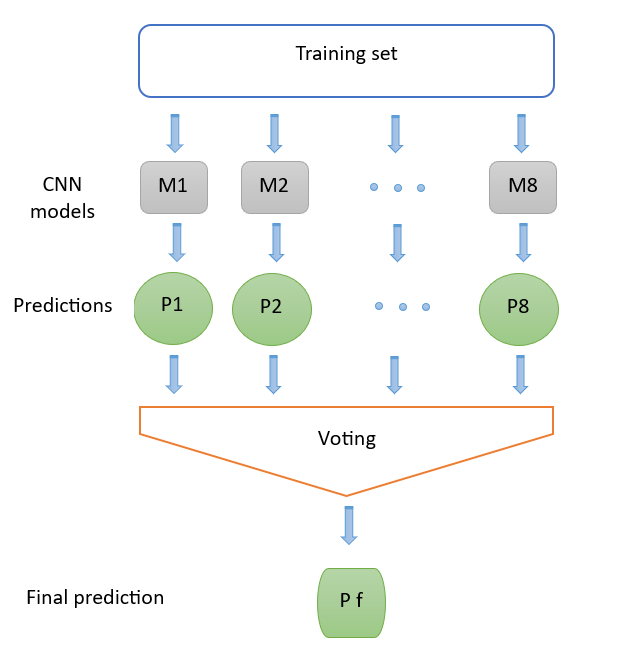}
\caption{Voting model architecture.}
\label{fig:vot}
\end{figure*}

%%%%%%%%%%%%%%%%%%%%%%%%
Several separate models are combined in ensemble learning to improve generalization performance. Deep learning architectures are now performing better than standard or shallow models. To improve the generalization performance of the final model, deep-enhanced learning models integrate the benefits of ensemble learning and deep learning. Using ensemble learning, architecture can operate effectively by combining its various parts to achieve a single objective. Numerous ensemble models exist, including voting, bagging, boosting, and stacking \cite{meel2021han}. Voting combines predictions from multiple models, making the overall system more robust; even if individual models fail or make errors, the ensemble can still provide reliable results, and it allows the combination of different model architectures or pre-trained embeddings to enhance the overall understanding of image content. In addition, ensemble methods such as voting reduce overfitting by averaging out model biases and generalizing better to unseen data \cite{articleJ} \cite{singh2022efficient}. We employ the voting approach to aggregate the predictions made by each technique. To obtain a more robust and accurate caption, the ensemble learning model trains multiple instances of the transformer with different random initializations or hyper-parameters and then combines their output through voting. The voting model is presented in Fig. \ref{fig:vot} to combine the results of each of the eight transformer models. The BLEU score-1 was considered for this purpose, and the prediction result will be accepted from the model that gains the highest BLEU score.

%\begin{figure*}[h]
%\centering
%\includegraphics[width=1\textwidth,height=\textheight,keepaspectratio]{all.png}
%\caption{Workflow of the proposed image captioning model}
%\label{fig:all}
%\end{figure*}

\section{Experimental results}
\label{Exp}
This section presents the results obtained from the proposed model and compares them with the latest models.

\subsection{Environment setup}
To assess the performance of the proposed model, a set of experiments was conducted using the Google Colab Pro+ framework, equipped with 52 GB of RAM and 1 TB of storage capacity for implementation purposes. The proposed model was trained with a batch size of 64, employing the Adam optimizer \cite{article1}, a learning rate set at 0.00001, 30 epochs with early stopping, and the ReLU activation function was utilized.

Our model employs a structured approach to training an image captioning system, focusing on effectively managing the learning rate. The loss function is defined using cross-entropy, which calculates the loss between predicted and true labels without reduction. To prevent overfitting, early stopping is implemented to monitor validation loss and halt training if no improvement is observed after a predetermined number of epochs, restoring the best model weights. A custom learning rate scheduler is utilized to dynamically adjust the learning rate throughout the training process. It starts at a low rate of 0.00001 and gradually increases, facilitating a stable training experience that enhances convergence and overall model performance. This scheduler works with the Adamax optimizer during model compilation, ensuring effective optimization while reducing the risk of overfitting. This approach promotes stable convergence during training and improves the model's ability to generalize to unseen data, which is essential for generating accurate and contextually relevant captions for images. The chosen parameters are based on empirical observations and established best practices in neural network training, with the aim of balancing efficient convergence and stable training while minimizing the risk of overfitting.

\subsection{Evaluation metrics}

%%%% just highlighted 
While direct human judgment is the simplest way to evaluate text generated for images, scalability is challenging due to nonreusable human effort and subjective nature. To overcome these challenges, various evaluation metrics assess the performance of image captioning systems. These metrics measure the systems' ability to generate linguistically acceptable and semantically valid phrases. However, the choice of the most significant metric depends on the specific objectives of the image captioning task. BLEU and ROUGE are often considered standard. However, recent research has shown the value of incorporating diverse metrics such as METEOR, CIDEr, and SPICE to provide a more comprehensive evaluation and performance results. The evaluation metrics applied in this study include BLEU, ROUGE, METEOR, CIDEr, and SPICE. Table\ref{tab:metrics} provides a summary of common assessment metrics in image captioning, while the following section discusses them in more detail.

\begin{table*}[htbp]
    \centering
    \caption{Performance assessment metrics in image captioning}
    \label{tab:metrics}
    \footnotesize
    \resizebox{\columnwidth}{!}{%
    \begin{tabular}{lll}
    \hline
    \textbf{Metric} & \textbf{Evaluation task} & \textbf{Methodology}  \\
    \hline
    BLEU\cite{papineni-etal-2002-bleu}       &  Machine translation  &  n-gram precision   \\ 
    ROUGE\cite{lin2004rouge} & Document summarization  &  	n-gram recall   \\ 
   METEOR \cite{banerjee2005meteor} &	Machine translation  & 	n-gram with synonym matching     \\   
    CIDEr \cite{vedantam2015cider} & Image captioning & tf-idf weighted n-gram similarity\\
     SPICE\cite{anderson2016spice}   & Image captioning & Scene-graph synonym matching    \\ 
     \hline
    \end{tabular}%
    }
\end{table*}

\subsubsection{Bilingual evaluation understudy (BLEU)}
BLEU is a metric that evaluates the quality of machine-generated text by comparing individual segments to a set of reference texts \cite{papineni-etal-2002-bleu}. Its approach varies with the number of references and the length of the text. BLEU scores are higher for short autogenerated text and range from 0 to 1. The comparisons of the gram and the bigram determine BLEU-1 and BLEU-2, with an empirically determined maximum order of four for optimal correlation with human judgments. BLEU assesses adequacy through unigram scores and fluency through higher n-gram scores. Although widely used and language-independent, BLEU has drawbacks. It favors brief output texts, and a high score does not guarantee higher quality, making it imperfect for certain evaluations \cite{10.1145/3617592}. 

\subsubsection{Recall-oriented understudy for gisting evaluation (ROUGE)}
ROUGE is a set of measures that evaluate text summaries by comparing word sequences and pairs to a database of human-written reference summaries \cite{inproceedings}. Originally designed for machine translation accuracy and fluency assessment, it quantifies sentence-level similarity using the longest common subsequence between candidate and reference sentences. Similarly to BLEU, ROUGE is also computed by varying the n-gram count. However, unlike BLEU, which is based on precision, ROUGE is based on recall values. It captures sentence-level structure with in-sequence word matches, allowing non-sequential matching. ROUGE-L is the version that is used in the evaluation of image and video captioning. It calculates the recall and precision scores of the longest common subsequences (LCS) between each generated sentence and its corresponding reference sentence.% ROUGE-1, ROUGE-2, ROUGE-W, and ROUGE-SU4 serve various evaluation tasks, and their metrics range from 0 to 1.

\subsubsection{Metric for explicit ordering translation evaluation (METEOR)}
METEOR is designed for machine translation evaluation and is considered more valuable than BLEU, with a stronger link to human evaluations \cite{banerjee2005meteor}. It calculates scores based on generalized unigram matches between a candidate sentence and human-written reference sentences. The precision, recall, and alignment of the matched words contribute to the score computation. In cases with multiple reference sentences, the candidate's final evaluation considers the best score among independently computed ones. METEOR considers unigram overlap and incorporates additional features like stemming and synonymy matching. It aims to address some limitations of BLEU and ROUGE by providing a more comprehensive evaluation \cite{10.1145/3617592}.

\subsubsection{Consensus-based image description evaluation (CIDEr)}
CIDEr is an image caption quality assessment paradigm that relies on human consensus \cite{vedantam2015cider}. Assesses the similarity of a generated sentence to a set of human-written ground-truth sentences. Using the TF-IDF weighting for each n-gram in the candidate phrase, CIDEr encodes their frequency in reference sentences. CIDEr evaluates the grammar, significance, and accuracy of image captions and descriptions. Unlike metrics that work with a limited number of captions per image, CIDEr employs consensus utilization, making it suitable for analyzing the agreement between generated captions and human assessments \cite{10.1145/3617592}. 

%%%%% just highlighted 
\subsubsection{Semantic propositional image caption evaluation (SPICE)}
SPICE is a semantic concept-based image caption evaluation metric based on semantic scene graphs \cite{anderson2016spice}. It uses a graph-based semantic representation extracted from image descriptions \cite{hossain2019comprehensive}. Generated and ground-truth captions are converted into an intermediate scene graph representation through semantic parsing to calculate the SPICE score. The F1 score derived from precision and recall measures the similarity between the generated and ground-truth caption scene graphs.
%%%%%%%%%%%%%%%%%%%%%%%

\subsection{Quantitative analysis}

%\subsubsection {Experimenting with Flicker8K}
Table \ref{tableFlickr8k} compares the results obtained by the proposed model with the latest methods and Fig. \ref{fig:correct}. As will be discussed soon, the proposed model exhibits superior performance, with the highest scores highlighted in bold. The result includes the research with models based on the Flickr8k dataset. The proposed model achieved the highest scores in BLEU-1, BLEU-2, and BLEU-3: 0.728, 0.495, and 0.323, respectively. These scores indicate how well our system’s predictions align with reference captions on $n$-gram overlap. Our model obtained the highest result of the METEOR score, 0.604, which evaluates the semantic similarity between the generated and reference captions. The SPICE score, which focuses on semantic content overlap, was used to assess the quality of image captions. Our model achieved the highest SPICE value of 0.164, indicating strong semantic alignment. We also get competitive results for ROUGE L (0.432) and CIDEr (0.604). ROUGE L measures the longest common subsequence between generated and reference captions, CIDEr considers word frequency and diversity.

%\subsubsection {Experimenting with Flicker30K}

To demonstrate the effectiveness of the suggested model, we contrasted its results with those of the most advanced models on Flickr30k datasets, as indicated in Table \ref{tab:tableFlickr30k} and Fig. \ref{fig:correct30k}. The suggested model outperformed the latest methods in Flickr30k datasets, as demonstrated by the table, as determined by the BLEU-1, BLEU-2, BLEU-3, and BLEU-4 scores. ROUGE L METEOR CIDEr showed that the model performs similarly to the state-of-the-art. It indicates that the proposed approach can provide meaningful and understandable human captions. The results derived from alternative metrics, including SPICE (0.387), confirm the efficacy of the model. It is important to remember that the majority of other methods do not share their results on this metric. SPICE scores provide a unique evaluation metric for image captioning by focusing on the semantic content of generated captions rather than just word overlap. In addition, SPICE scores correlate better with human judgments of caption quality, making them a more reliable measure of how well captions align with human expectations.

% \usepackage{graphicx}
\begin{table*}[htbp]
   \centering
   \caption{Comparison of Flicker8K dataset image captioning}
   \label{tableFlickr8k}
   \resizebox{\columnwidth}{!}{%
   \begin{tabular}{lllllllll}
   \hline
   \textbf{Reference} & \textbf{B1\(\uparrow\)}   & \textbf{B2\(\uparrow\)}   & \textbf{B3\(\uparrow\)} & \textbf{B4\(\uparrow\)}   & \textbf{ROUGE L\(\uparrow\)} & \textbf{METEOR\(\uparrow\)} & \textbf{CIDEr\(\uparrow\)} & \textbf{SPICE\(\uparrow\)} \\ \hline

%Deep Visual-Semantic Alignments for Generating Image Descriptions%
   \cite{karpathy2015deep} Karpathy et al.(2015) & 0.579          & 0.383          & 0.245        & 0.160          & NA                & NA               & NA              &NA              \\ \hline

%Modeling coverage with semantic embedding for image caption generation %
  \cite{jiang2019modeling} Jiang et al.(2019) & 0.690	& 0.471	& 0.324	& 0.219	&{ 0.502}	& 0.203	& 0.507	& NA
   \\ \hline

%Hyperparameter analysis for image captionin %
   \cite{patel2020hyperparameter} Patel et al.(2020) & 0.601	&0.414	&0.274	&0.181	&0.433	&0.183	&0.452 & NA
   \\ \hline

    \cite{katpally2020ensemble} Katpally et al.(2020) & 0.634	&0.400	&0.287	&0.151 & NA  & NA & NA  & NA  \\ \hline
   
%Image Caption Generation Using CNN-LSTM Based  Approach%
  \cite{bineeshia2021image} Bineeshia et al.(2021) & 0.589          & 0.335          & 0.263        & 0.148          & NA                & NA               & NA              & NA              \\ \hline

%Image Caption Generator Using Convolutional Recurrent Neural Network Feature Fusion%
   \cite{dahriimage} Dahri et al.(2023) & 0.603    & 0.360          & 0.220   & 0.122          & NA                &NA               & NA              & NA  \\ \hline
   
%Towards local visual modeling for image captioning %
   \cite{ma2023towards} Ma et al.(2023) & 0.674	&NA &NA	& 0.243	& 0.448	& 0.215	& {0.636} &NA
   \\ \hline

   \textbf{The Proposed Model} & \textbf{0.728} & \textbf{0.495} & \textbf{0.323}        & 0.208          & 0.432  &\textbf{0.235 }          &{0.604} & \textbf{0.164} \\ \hline
   \end{tabular}%
   }
\end{table*}

\begin{table*}[htbp]
   \centering
   \caption{Comparison of Flicker30K dataset 
    image captioning}
   \label{tab:tableFlickr30k}
   \resizebox{\columnwidth}{!}
   {%
   \begin{tabular}{lllllllll}
   \hline
   \textbf{Reference}& \textbf{B1\(\uparrow\)}   & \textbf{B2\(\uparrow\)}   & \textbf{B3\(\uparrow\)} & \textbf{B4\(\uparrow\)}   & \textbf{ROUGE L\(\uparrow\)} & \textbf{METEOR\(\uparrow\)} & \textbf{CIDEr\(\uparrow\)} & \textbf{SPICE\(\uparrow\)} \\ \hline
   \cite{karpathy2015deep} Karpathy et al.(2015) & 0.573  & 0.369  & 0.240 & 0.157 & NA   & NA  & NA  & NA            \\ \hline

   \cite{7780872} You et al.(2016) &0.647 	&0.460 	&0.324 	&0.230 	&0.189 & NA  & NA & NA           \\ \hline
   
%Aligning Where to See and What to Tell: Image Captioning with Region-based Attention and Scene-specific Contexts

   \cite{fu2016aligning} Fu et al.(2016) &0.649 &0.462 &0.324 &0.224 &0.451 &0.194  &0.472 &NA \\ \hline
   
   \cite{lu2017knowing} Lu et al.(2017) & 0.677 & 0.494 &0.354 &0.251 &0.204  & NA &0.531 & NA \\ \hline

%Image captioning with text-based visual attention % 
   \cite{he2019image} He et al.(2019) & 0.666	& 0.484	& 0.346	& 0.247	& 0.467	& 0.202	& 0.524	& NA
   \\ \hline
% Modeling coverage with semantic embedding for image caption generation% 
    \cite{jiang2019modeling} Jiang et al.(2019) & 0.689	& 0.468	&0.319	&0.220	&0.487	&0.191	&0.428	& NA
   \\ \hline

%Reference-based model using multimodal gated recurrent units for image captioning%
   \cite{do2020reference} Do et al.(2020) &0.695	&0.463	&0.341	&0.232	&0.451	&0.302 &	0.486 & NA\\ \hline

% Fusion models for improved image captioning%
   \cite{kalimuthu2021fusion} Kalimuthu et al.(2021)  & 0.647	& 0.456	& 0.320	& 0.224	& 0.449	& 0.197	& 0.467	& 0.136
   \\ \hline  

%   \color{teal}\cite{rodoshi2023automated} Rodosh et al.(2023)\color{black}  &0.550	&0.338	&0.237	&0.119 & NA & NA & NA & NA\\ \hline

   %Towards local visual modeling for image captioning %
   \cite{ma2023towards} Ma et al.(2023) & 0.671	& NA & NA		& 0.233	& 0.443	& 0.204	& 0.645 & NA
   \\ \hline
   
% NumCap: A Number-controlled Multi-caption Image Captioning Network%
   \cite{abdussalam2023numcap} Abdussalam et al.(2023) & 0.694	& 0.498	& 0.355	& 0.254	& 0.538	& 0.251	& 0.469	&NA
   \\ \hline
   \textbf{The Proposed Model} & \textbf{0.798} & \textbf{0.561} & \textbf{0.387} &\textbf{0.269}  & 0.443 & {0.213}   & 0.565 & \textbf{0.387} \\ \hline
   \end{tabular}%
   }
\end{table*}

\subsection{Qualitative analysis}

\begin{figure*}[ht]
\centering
\includegraphics[width=10cm]{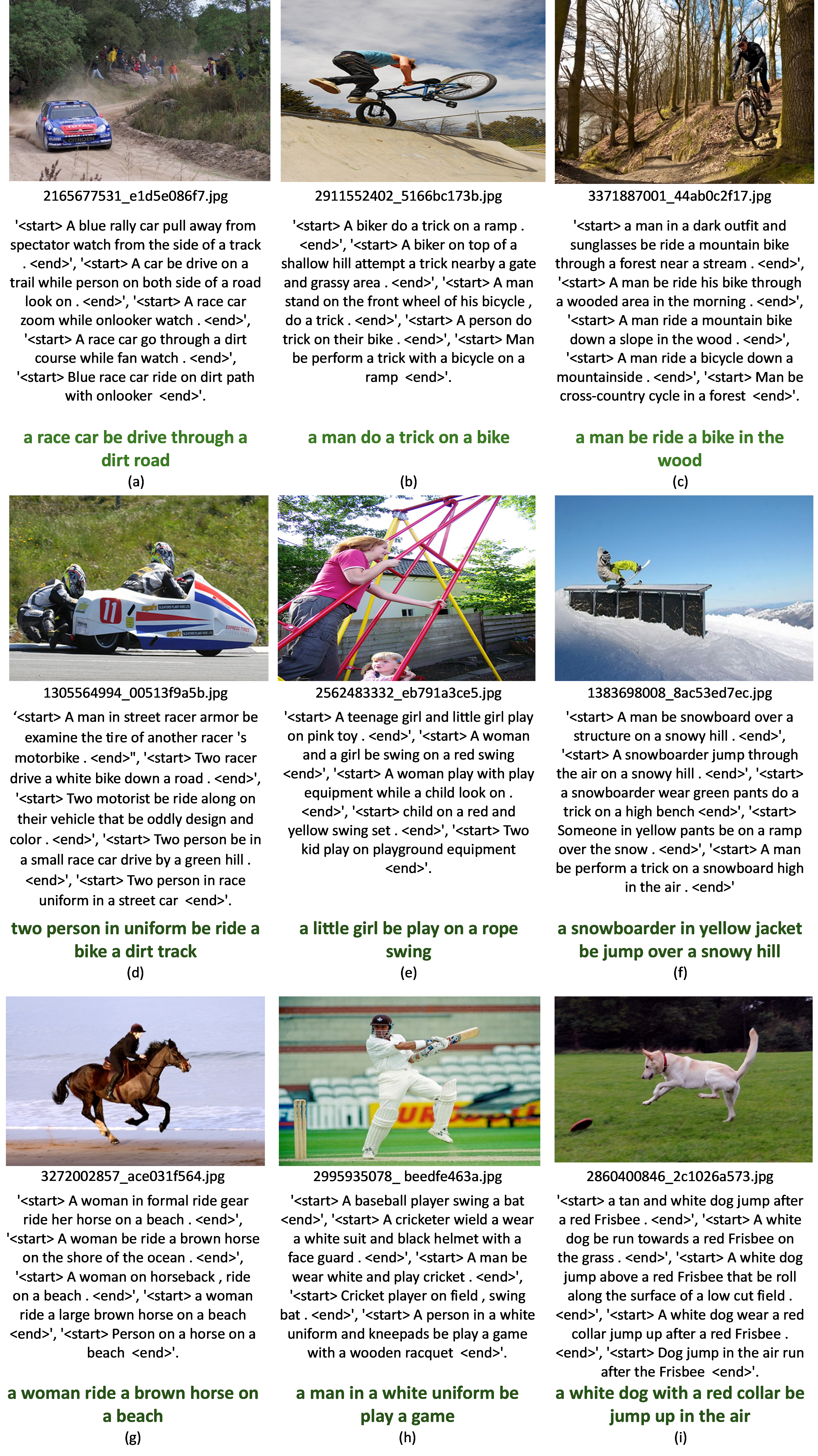}
\caption{Samples of correct captions (green text) generated   by the proposed model from Flicker8K dataset}
\label{fig:correct}
\end{figure*}

\begin{figure*}[ht]
\centering
\includegraphics[width=9.5cm]{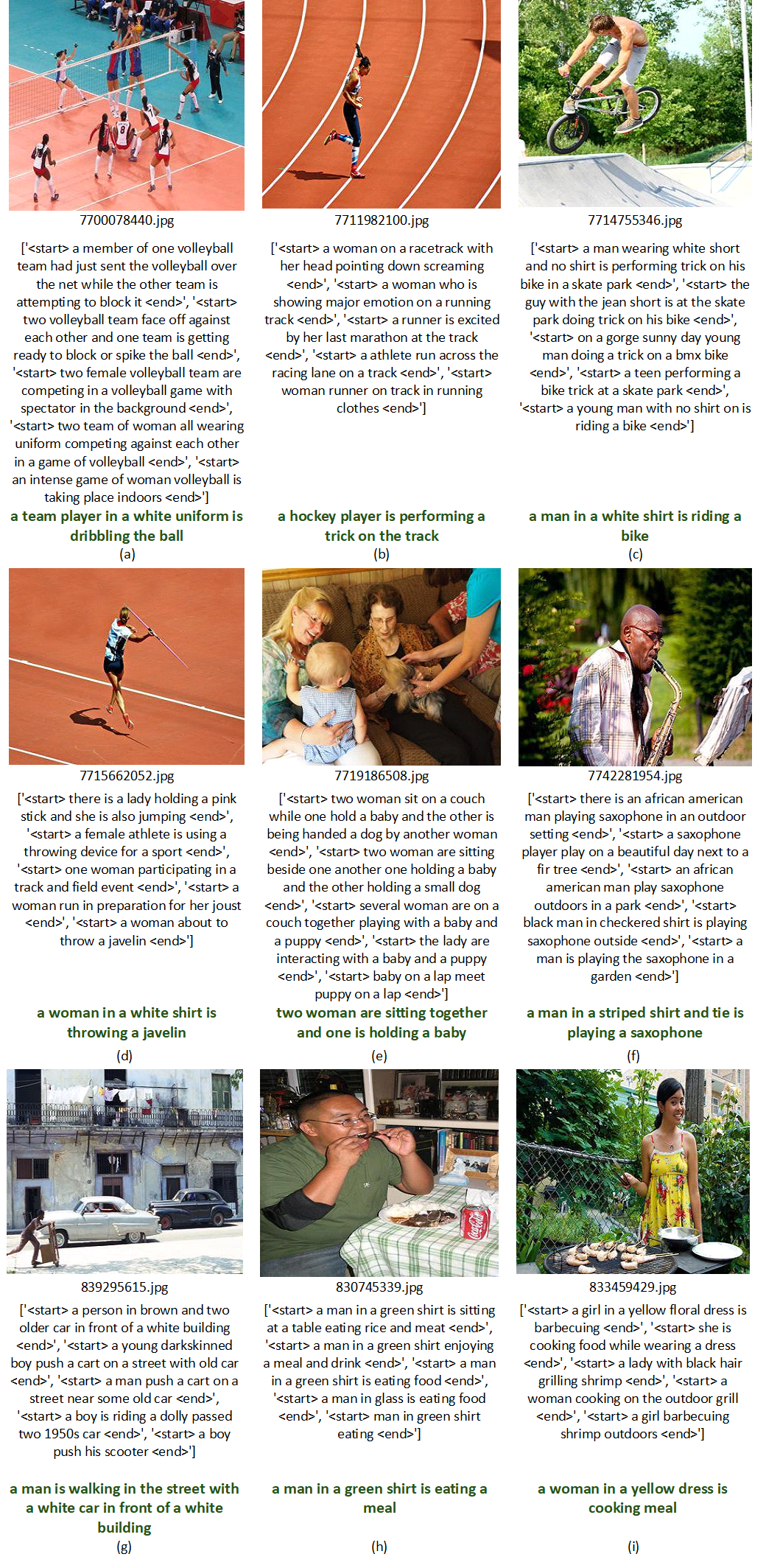}
\caption{Samples of correct captions (green text) generated by the proposed model from Flicker30K dataset}
\label{fig:correct30k}
\end{figure*}

As demonstrated in Fig. \ref{fig:correct} and Fig. \ref{fig:correct30k}, we have provided several sentences produced by our caption method to validate the effectiveness of our model.
In general, our model demonstrates proficiency in generating captions that are not only relevant but also accurate in describing the image content. Fig. \ref{fig:correct} presents samples of nearly correct captions from the Flicker8K dataset. Green text is used to identify the generated captions. As noted in Fig. \ref{fig:correct}.b, where the man is not riding the bike in the position in the picture, it is revealed that he is performing a trick by ``do the trick''. In Fig. \ref{fig:correct}.c, despite the terms ``forest'' and ``wood'' appearing in the references, the model was able to accurately depict the appearance of wood in the image accurately; however, since the image is not a green forest, the description produced by the system used the more correct word ``wood''. The description of the sites was available through the picture Fig. \ref{fig:correct} .a in the sentence ``dirt road'', through the second picture Fig. \ref{fig:correct}.f in the sentence “over a snowy hill“, and Fig. \ref{fig:correct}.g on a beach. The ``uniform'' was generated for both Fig. \ref{fig:correct}. d and Fig. \ref{fig:correct}.h  to clarify that they belong to a certain outfit.

On the other hand, Fig.\ref{fig:correct30k} shows the correct samples from Flicker30k. It is clear in Fig.\ref{fig:correct30k}.e, how the number of women was accurately identified for the first scenario, particularly in settings with complex backgrounds, counting the number of items is a higher level of artificial intelligence than object recognition \cite{jiang2019modeling}, we can observe how the model determines the gender, (man) in figures: Fig.\ref{fig:correct30k}.c, Fig.\ref{fig:correct30k}.d, Fig.\ref{fig:correct30k}.h, and Fig.\ref{fig:correct30k}.g, and (woman) in Fig.\ref{fig:correct30k}.b, Fig.\ref{fig:correct30k}.e, and Fig.\ref{fig:correct30k}.i. An illustration of how the model can represent the location of an object as ``in front of'' is provided in Fig.\ref{fig:correct30k}.g.  Furthermore, the example in Fig.\ref{fig:correct30k}.b ``is performing a trick'' and Fig.\ref{fig:correct30k}.a ``in a white uniform'' effectively conveys the setting.  ``Is eating'' in Fig.\ref{fig:correct30k}.h and ``Is cooking'' in Fig.\ref{fig:correct30k}.i serve as an illustration of how to differentiate between the actions related to a meal. 
Objects such as ``saxophone'' Fig.\ref{fig:correct30k}.f and ``javelin'' Fig.\ref{fig:correct30k}.d were correctly recognized. However, in Fig.\ref{fig:correct30k}.c it is evident that the model interprets the white area as the color of the shirt. A single thing might have several characteristics depending on the situation at hand. Learning to identify attributes in computer vision is still a challenging task \citep{jiang2019modeling}.

\begin{figure}%[ht]
\centering
\includegraphics[width=0.9\textwidth,height=\textheight,keepaspectratio]{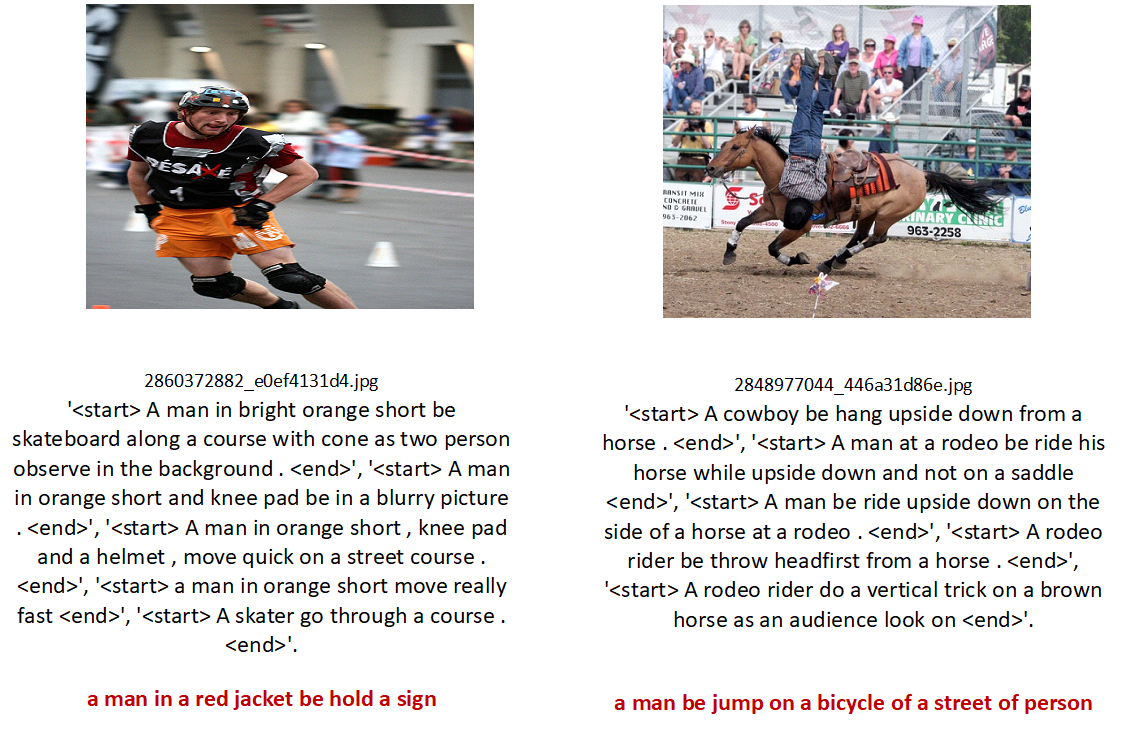}
\caption{Samples of incorrect captions (red text) generated   by the proposed model from Flicker8K dataset}
\label{fig:incorrect}
\end{figure}

\begin{figure}%[ht]
\centering
\includegraphics[width=0.9\textwidth,height=\textheight,keepaspectratio] {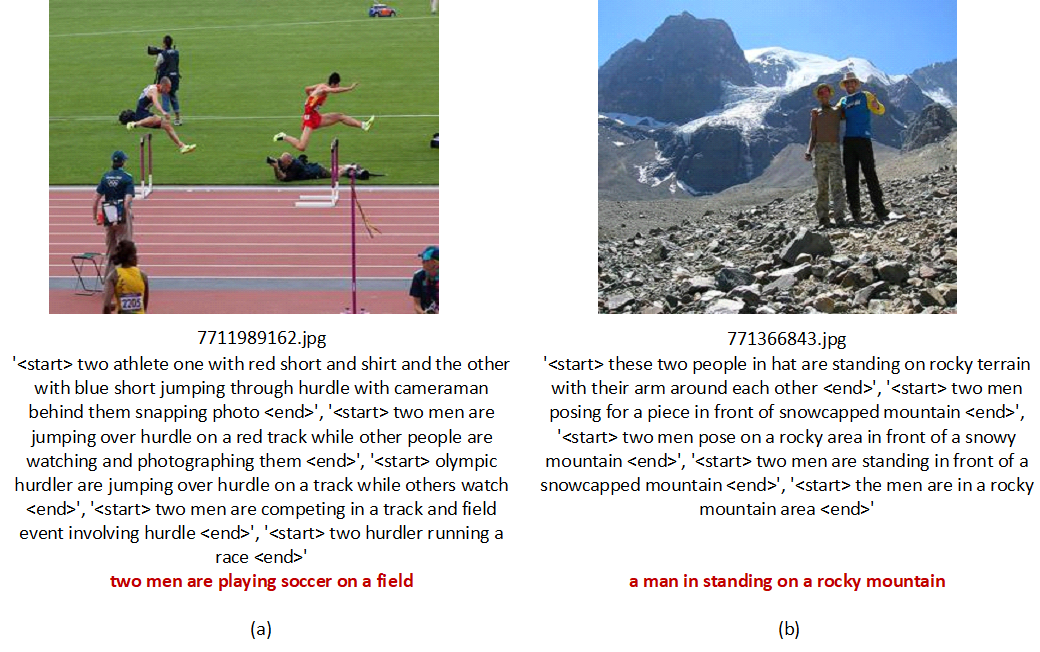}
\caption{Samples of incorrect captions (red text) generated by the proposed model from Flicker30K dataset}
\label{fig:incorrect30k}
\end{figure}

However, incorrect captions are shown in Fig.\ref{fig:incorrect} from the Flicker8K dataset. The right figure mistakenly describes the signs in the man's shirt as ``hold a sign'', whereas the left figure was incorrectly described. There are errors in the generated caption of the Flicker30k dataset, examples displayed in Fig.\ref{fig:incorrect30k}, and the created captions are identified by red text. The left figure (the obstacle on a red track) was incorrectly described as ``soccer on a field.'' However, the number of players was correctly listed as ``two''. In the right figure, the model produced the place in the caption ``standing on a rocky mountain,'' while the number (two men) was incorrectly provided as ``a man.'' These inaccurate captions show how the existing image captioning model has difficulty recognizing actions, context, and complicated settings. The model may fail to recognize context, leading to erroneous interpretations. For future plans, it may be necessary to detect and distinguish items within images effectively; object recognition algorithms should be improved. Improved context analysis methods should also be used to understand the connections between actions and objects.

\subsection{Ablation study}

An ablation study was conducted to assess the contributions of individual components within the proposed ensemble model for image captioning. The Flicker8K dataset was utilized for this purpose. In this study, evaluation metrics such as BLEU, ROUGE, METEOR, CIDEr, and SPICE were applied. This section outlines the methodology used and presents the findings, highlighting the importance of each model within the architecture. The following ensemble configurations were analyzed:\\

\begin{enumerate}
    \item \textbf{Baseline 1:} MobileNetV2, VGG16, VGG19, and ResNet50. These CNN models were used for feature extraction, capturing essential visual elements from the images before the transformer processes the extracted features for text generation.
    \item \textbf{Baseline 2:} MobileNetV2, VGG16, VGG19, ResNet50, and ResNet101. Like Baseline 1, this configuration incorporates additional CNNs to enhance feature extraction, providing a richer representation for the transformer during caption generation.
    \item \textbf{Baseline 3:} MobileNetV2, VGG16,VGG19, ResNet50, ResNet101, RegNetX120 and EfficientNetB4. This ensemble combines multiple CNN models to maximize feature extraction capabilities, allowing the transformer to generate more accurate and contextually relevant captions.
    \item\textbf{Full Ensemble Model:} In the full model, we evaluated the performance of the complete ensemble model, which integrates the outputs of all CNNs and a transformer language model to generate captions. 

\end{enumerate}
 
The results are summarized and compared in Table \ref{tab:ensemble}.

% \usepackage{booktabs}
% \usepackage{graphicx}

\begin{table*}[htbp]
   \centering
   \caption{Comparison of ensemble models using Flicker8K dataset}
   \label{tab:ensemble}
   \resizebox{\columnwidth}{!}
   {%
  \begin{tabular}{@{}lllllllll@{}}
  \toprule
  \textbf{Model}               & \textbf{B1\(\uparrow\)}    & \textbf{B2\(\uparrow\)}    & \textbf{B3\(\uparrow\)}    & \textbf{B4\(\uparrow\)}    & \textbf{ROUGE L\(\uparrow\)} & \textbf{METEOR\(\uparrow\)} & \textbf{CIDEr\(\uparrow\)} & \textbf{SPICE\(\uparrow\)} \\ \midrule
  \textbf{Baseline 1}          & 0.656          & 0.428          & 0.269          & 0.165          & 0.395            & 0.208           & 0.446          & 0.132          \\
  \textbf{Baseline 2}          & 0.697          & 0.470          & 0.304          & 0.192          & 0.419            & 0.226           & 0.544          & 0.154          \\
  \textbf{Baseline 3}          & 0.705          & 0.476          & 0.309          & 0.197          & 0.421            & 0.227           & 0.545          & 0.156          \\
  \textbf{Full ensemble model} & \textbf{0.728} & \textbf{0.495} & \textbf{0.323} & \textbf{0.208} & \textbf{0.432}   & \textbf{0.235}  & \textbf{0.604} & \textbf{0.164} \\ \bottomrule
  \end{tabular}%
  }

\end{table*}

The results indicate that the ensemble model consistently outperforms each base model in all metrics, validating the hypothesis that the combination of multiple models enhances feature representation. The complete ensemble demonstrated the highest scores on all metrics, indicating that the diversity of features captured by multiple CNNs leads to improved caption quality. This ensemble approach mitigates the limitations inherent in individual models by leveraging their unique strengths.

%% just highlighted

\subsection{Discussion}
The results demonstrated how our model differs from other methods in feature text extraction by focusing on salient image regions and characteristics through attention mechanisms. Table \ref{tableFlickr8k} and Table \ref{tab:tableFlickr30k} highlight the important distinctions between our suggested model and the other models and emphasize the research contributions in the following areas:(1) enhanced prediction robustness: In contrast to previous methods, our model uses an ensemble learning strategy, which effectively combines eight CNN models via a voting process, to fine-tune the ideal caption for every image. This increases the architecture's robustness and generalizability, while greatly improving its efficiency. Our model efficiently reduces overfitting by combining predictions from many base models, resulting in a more robust and flexible solution. (2) comprehensive evaluation metrics: to gain a deeper understanding of the model's capabilities, we used a methodology in this research work that took a variety of indicators into account. A more realistic description of the overall performance of a recently suggested model in this growing field will come from a comprehensive evaluation that considers multiple factors.

%%%%%%%%%% EiC 
To further validate our findings, we conducted paired t-tests between the full
ensemble and each ablation variant. The results revealed that the performance
differences were statistically significant (p $<$ 0.05), reinforcing the necessity of
using an ensemble model. 

Several models continue to show difficulties, such as explosions in gradients and inaccurate sentence construction, which impact the image encoding and description process. Most modern captioning models use LSTM and RNNs as language models. However, long-term information must go through every cell before arriving at the present processing cell, since RNN and LSTM operate sequentially. As a result, it is readily tainted by repeatedly multiplying it by small values smaller than zero, leading to vanishing gradients that delay updating the network weights and the learning process. Some, but not all, of the issues with the disappearing gradient can be resolved via LSTM. Moreover, LSTM and RNN require additional resources and are not hardware-friendly. While LSTM-based image descriptions yield remarkable results, putting them into images requires more time, work, and parameters. An innovative caption creation framework was introduced by \citep{yang2020ensemble}, the EnsCaption framework, combining caption generation and retrieval with a re-ranking procedure and adversarial network for improved accuracy. While EnsCaption shows strong performance it has limitation in recognition of fine-grained features.
%%%%%%%%%% R3

Using an ensemble of CNN models in our image captioning framework significantly enhances performance, improving accuracy, and robustness. Ensembles generalize more effectively to unseen data, which is crucial for applications requiring high model reliability. This is especially significant in domains with high variability, such as medical imaging or environmental monitoring. However, this enhancement increases computational demands, leading to longer training times. In critical fields such as medical imaging, even a slight improvement in accuracy can impact patient outcomes, as improved detection rates can reduce false negatives and ensure timely treatment. The significance of these trade-offs is also evident in security applications, where accurate image analysis improves resource allocation, and in disaster response, where precise facial recognition can identify potential threats. Therefore, despite the considerable computational requirements, the significant advantages in essential applications justify adopting ensemble methods.

\subsection{Real-World applications of the proposed image captioning model}

%Examples of how 
Our proposed image captioning model could be applied in real-world scenarios in enhanced search engines, a search engine integrates an image captioning model to improve the search experience by delivering more informative and relevant image results. When a user enters a search query for images, the model analyzes each image in the database and generates detailed captions that accurately describe the content. These captions are indexed alongside the images, enabling the search engine to retrieve results that align more closely with the user's query. %This enhancement significantly improves the overall search experience, making it easier and quicker for users to find what they want.
Another compelling use case is assistive technology for the visually impaired. 
%where an image captioning system enhances users' understanding of their surroundings. 
In this scenario, visually impaired individuals use a mobile application to capture photos of their environment. The image captioning system analyzes these images and generates descriptive audio captions that highlight key elements. %, such as objects, people, and activities within the scene.  
The application significantly improves the user's quality of life by facilitating greater interaction with their surroundings.

\section{Limitation and future work}

The proposed Attention-Based Transformer Model for Image Captioning encounters several challenges. First, the model sometimes misinterprets the color of certain areas as corresponding to different areas or clothing items, highlighting the difficulty of recognizing multiple attributes associated with a single factor in computer vision. Second, there are instances where the model does not accurately count the number of elements in the target image; this task involves a higher level of artificial intelligence than simple object recognition. In addition, the model may struggle to understand complex settings, leading to incorrect interpretations. Lastly, a key limitation that affects model performance in image captioning is the presence of noisy or ambiguous images. Although this issue falls outside the scope of the current research, it is important to note that such images, characterized by distracting elements, low resolution, or unclear subjects, can hinder the model's ability to accurately interpret visual content, resulting in incorrect or irrelevant captions.

Future advances in image captioning can effectively address existing limitations through focused research initiatives. First, improving object recognition algorithms will enhance the model's ability to detect and distinguish items within images accurately. Second, improving the robustness of the model against noisy or ambiguous input to improve the quality of the caption and the overall performance in various real-world scenarios. Developing more comprehensive evaluation metrics can also offer a deeper assessment of caption quality. Furthermore, exploring how to handle diverse datasets effectively, ensuring that the model can generalize well across different scenarios. Finally, enhancing the visualization of image captioning models will allow insight into how the model focuses on specific areas of an image when generating captions.
\section{Conclusions}

Converting an input image into an explanation in words is known as image captioning. It can be used in various situations, including social networking, smart travel, assisting the blind, medical image captioning for healthcare, education, and training of children. Competition among researchers is leading to an increase in the number of unique models.
In this research, we thoroughly investigated the transformer model and provided many helpful ways to attract attention. We have shown the potential for significant advancement in this field by implementing an attention mechanism in a transformer-based design. We introduce a novel ensemble learning framework to generate captions based on a voting mechanism that selects the caption with the highest bilingual evaluation understudy (BLEU) score. This framework enhances the richness of the captions generated by utilizing multiple deep neural network architectures. The robustness and efficacy of the proposed approach have been demonstrated by a comprehensive analysis of the Flickr8K and Flickr30K datasets combined with common metrics and measurements. We believe that our approach will encourage future research to explore transformer-based designs further and push the limits of what is possible in multilingual image captioning.

\section{Declaration}
%\textbf{Funding:}
%No funding is available.

%\hfill \break\noindent\textbf{Conflict of Interest:} 
%The authors declare that they have no known competing financial interests or personal relations that could have appeared to influence the work reported in this paper. 

\hfill \break\noindent\textbf{Data Availability:} 
The datasets used in this study are publicly available and can be downloaded by the following links: Flickr8k (https://www.kaggle.com/datasets/adityajn105/flickr8k), Flickr30K https://www.kaggle.com/datasets/eeshawn/flickr30k)

%By publishing our research, we intend to generate cooperative efforts and creative solutions that will advance models' understanding and description of visual content, the research that has been conducted has wider implications for the interaction of computer vision and natural language processing.
%%%%%%%%%%%%%%%%%%%%%%%%%%%%%%%%%%%%%%%%%%%%%%%%%%%%%%%%%%%%%%%%%%%%%%%%%%%%%%%%%%%%%%%%%%%%%%%%%%%%%%%%%%%%%%%%%%%%%%%%%%%%%%%%%%%%%%%%%%%%%%%%%%%%%%%%%%%%%%%%%%%%%%%%%

     
\bibliography{sn-bibliography}% common bib file

\end{document}